\documentclass[final]{cvpr}

\usepackage{times}
\usepackage{epsfig}
\usepackage{graphicx}
\usepackage{amsmath}
\usepackage{amssymb}
\usepackage{enumitem}

\usepackage[pagebackref=true,breaklinks=true,colorlinks,bookmarks=false]{hyperref}

\renewcommand*{\ie}{i.e.\@\xspace}
\renewcommand*{\eg}{e.g.\@\xspace}

\newcommand{\Expect}{{\rm I\kern-.3em E}}

\def\cvprPaperID{7246} % *** Enter the CVPR Paper ID here
\def\confYear{CVPR 2021}
\begin{document}

%%%%%%%%% TITLE
\title{Benchmarking Representation Learning for Natural World Image Collections}

\author{\hspace{-5pt}Grant Van Horn$^{1}$\hspace{10pt}Elijah Cole$^{2}$\hspace{7pt}Sara Beery$^{2}$\hspace{7pt}Kimberly Wilber$^{3}$\hspace{7pt}Serge Belongie$^{1,3}$\hspace{7pt}Oisin Mac Aodha$^{4}$\\$^{1}$Cornell University \hspace{30pt} $^{2}$Caltech   \hspace{30pt}$^{3}$Google\hspace{30pt}  $^{4}$University of Edinburgh \\\url{www.github.com/visipedia/newt}  }

\maketitle

%%%%%%%%% ABSTRACT
\begin{abstract}
Recent progress in self-supervised learning has resulted in models that are capable of extracting rich representations from image collections without requiring any explicit label supervision.  
However, to date the vast majority of these approaches have restricted themselves to training on standard benchmark datasets such as ImageNet. 
We argue that fine-grained visual categorization problems, such as plant and animal species classification, provide an informative testbed for self-supervised learning. 
In order to facilitate progress in this area we present two new natural world visual classification datasets, iNat2021 and NeWT. 
The former consists of 2.7M images from 10k different species uploaded by users of the citizen science application iNaturalist. 
We designed the latter, NeWT, in collaboration with domain experts with the aim of benchmarking the performance of representation learning algorithms on a suite of challenging natural world binary classification tasks that go beyond standard species classification. 
These two new datasets allow us to explore questions related to large-scale representation and transfer learning in the context of fine-grained categories. 
We provide a comprehensive analysis of feature extractors trained with and without supervision on ImageNet and iNat2021, shedding light on the strengths and weaknesses of different learned features across a diverse set of tasks. 
We find that features produced by standard supervised methods still outperform those produced by self-supervised approaches such as SimCLR. 
However, improved self-supervised learning methods are constantly being released and the iNat2021 and NeWT datasets are a valuable resource for tracking their progress.  
\end{abstract}

%%%%%%%%% BODY TEXT
\section{Introduction}
\begin{figure}[t!]
\centering
\includegraphics[width=0.47\textwidth]{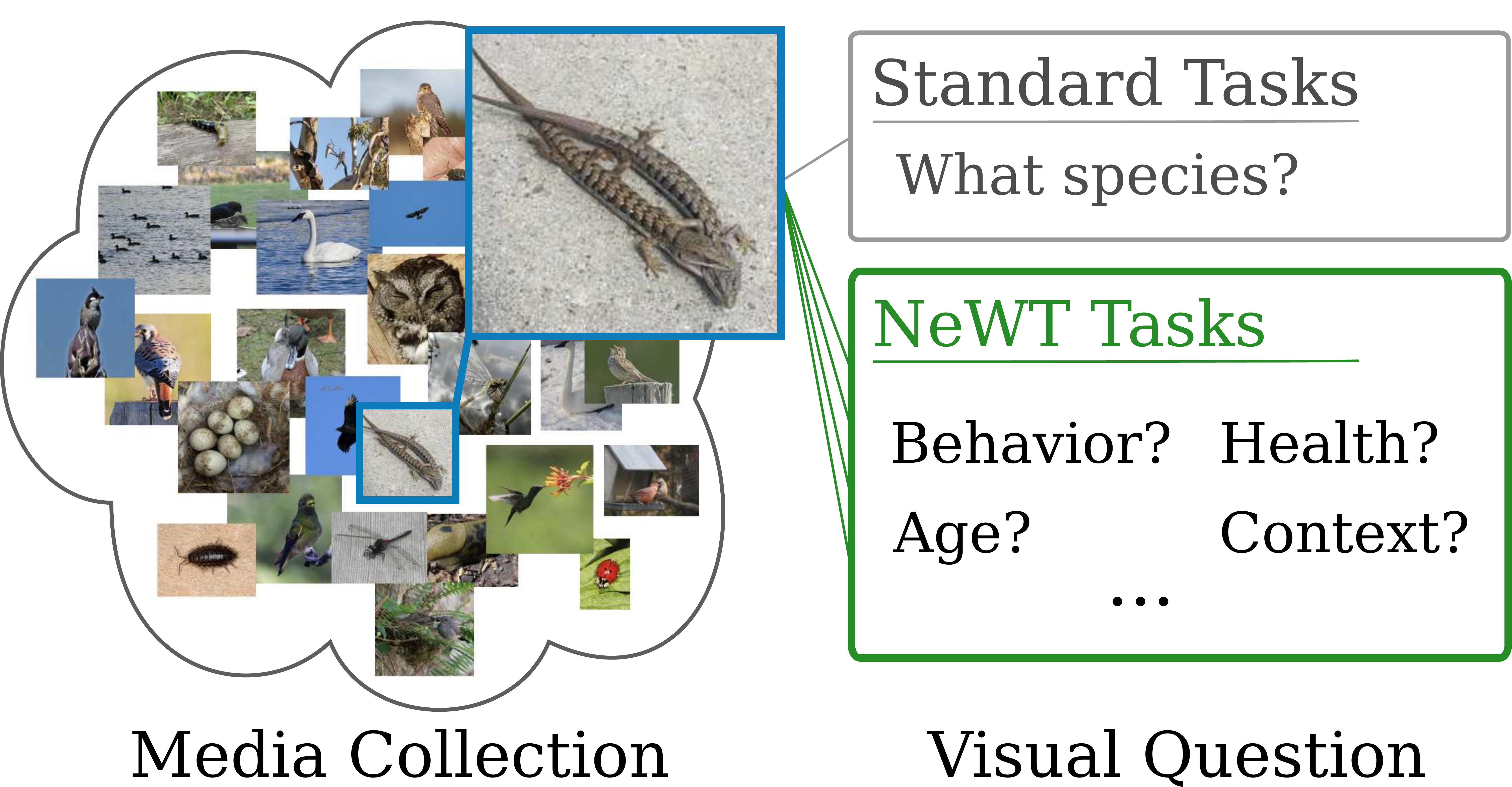} 
\vspace{5pt}
\caption{Existing fine-grained image datasets are typically focused on a single task \eg species identification. As natural world media collections grow, we have the opportunity to extract information beyond species labels to answer important ecological questions. 
For example, with the help of community scientists, researchers from the NHMLA were able to curate over 500 images of alligator lizards mating, a phenomenon seldomly recorded in the existing scientific literature~\cite{giaimo_NYT_lizard}. 
We analyze if trained feature extractors can answer similar novel image understanding questions with minimal additional training and present \textbf{NeWT}, a diverse benchmark of natural world visual understanding tasks such as animal health, life-stage, behavior, among others.
}
\vspace{-5pt}
\label{fig:teaser}
\end{figure}

Learning representations of images through self-supervision alone has seen impressive advancement over the last few years. 
There are tantalizing results that show self-supervised methods, fine-tuned with $1\%$ of the training labels, reaching the performance of their fully supervised counterparts~\cite{chen2020big}. 
In many domains, aggregating large amounts of data is typically not the bottleneck. 
Rather, it is the subsequent labeling of that data that consumes vast amounts of money and time. 
This is further compounded in fine-grained domains, \eg medicine or the natural world, where sufficiently well trained annotators are few or their time is expensive. 
If the benefits of self-supervised learning come to full fruition,
then the applicability and impact of computer vision models across many domains will see a rapid increase.

One particular domain that is well suited for this type of advancement is the study of the natural world through photographs collected by communities of  enthusiasts. 
Websites such as iNaturalist~\cite{iNatWeb} and eBird~\cite{sullivan2009ebird} amass large collections of media annually. 
To date, there are 60M images in iNaturalist spanning the tree of life and 25M images of birds from around the world in eBird, both representing point-in-time records of wildlife. 
Identifying the species in an image has been well studied by the computer vision community \cite{wah2011caltech,KhoslaYaoJayadevaprakashFeiFei_FGVC2011,berg2014birdsnap,gvanhorn2018inat}, however this is only the tip of the iceberg in terms of questions one may wish to answer using these vast collections.
These datasets contain evidence of the health and state of the individuals depicted, along with their behavior. 
Having an automated system mine this data for these types of properties could help scientists fill in missing pieces of basic natural history information that are crucial for our understanding of global biodiversity and help measure the loss of biodiversity due to human impact~\cite{cardinale2012biodiversity}. 

To give one example, science is ignorant to the nesting requirements of thousands of bird species, including the vulnerable Pink-throated Brilliant (Heliodoxa gularis) \cite{pinkthroatedbrilliant}. 
Knowing how and where this species builds its nest is a crucial piece of information needed when discussing conservation based interventions, particularly as it pertains to the ability of this species to exist in degraded and fragmented habitats~\cite{pinkthroatedbrilliant}. 
While nothing can replace the capabilities of a biologist in the field, citizen science projects like eBird and iNaturalist are collecting raw images that could help answer some of these questions. 
However, herein lies the problem. 
It is currently a daunting task to label  training datasets for these specialized questions that would satisfy the data appetite of an off-the-shelf deep network.

Self-supervised learning is one potential solution that could alleviate the labeling burden by taking advantage of large media collections. While most research on self-supervised learning focuses on ImageNet~\cite{russakovsky2015imagenet}, in this work we expand these techniques to the natural world domain and fine-grained classification.
Following Goyal~\etal~\cite{goyal2019scaling}, we maintain that a good representation should generalize to many different tasks, with limited supervision or fine-tuning.
We do not investigate self-supervised learning as an initialization scheme for a model that is further optimized and finetuned, but rather as a way to learn feature representations themselves. 
Importantly, \cite{goyal2019scaling} point out that self-supervised feature learning and subsequent feature evaluation on the same dataset does not test the generalization of the features. 
Inspired by this, we present a new large-scale pretraining dataset and new benchmark tasks specifically designed to enable us to ask questions about the generalization of self-supervised learning on natural world image collections.

We make the following three contributions:
%\vspace{-3pt}
\begin{itemize}[topsep=0pt,itemsep=-3pt,partopsep=1ex,parsep=1ex]
    \item \textbf{iNat2021} - A new large-scale image dataset collected and annotated by community scientists that contains over 2.7M images from 10k different species.
    \item \textbf{NeWT} - A new suite of 164 challenging natural world visual benchmark tasks that are motivated by real world image understanding use cases. 
    \item A detailed evaluation of self-supervised learning in the context of natural world image collections. 
    We show that despite recent progress, self-supervised features still lag behind supervised variants.   
\end{itemize}

\section{Related Work}
\subsection{Learning Visual Representations} 

Transfer learning using features extracted from deep networks that have been trained via supervision on large datasets results in powerful features that can be applied to many downstream tasks~\cite{donahue2014decaf,yosinski2014transferable}. 
However, there is evidence to suggest that pretraining on datasets such as ImageNet~\cite{russakovsky2015imagenet} is less effective on fine-grained categories when the labels are not well represented in the source dataset~\cite{kornblith2019better}. 
Self-supervised learning, \ie learning visual representations without requiring explicit label supervision, is an exciting research area that, if successful, could provide a much more scalable way to learn representations for a wide variety of tasks -- including fine-grained ones. 

Earlier work in self-supervised learning in vision involved framing the learning problem via proxy tasks \eg predicting context from image patches~\cite{doersch2015unsupervised,noroozi2016unsupervised}, image colorization~\cite{zhang2016colorful}, or predicting image rotation~\cite{gidaris2018unsupervised}, to name a few. 
The most effective recent approaches have focused on contrastive learning based training objectives~\cite{hadsell2006dimensionality,gutmann2010noise}, where the aim is to learn features from images such that augmented versions of the same image are nearby in the feature space, and other images are further away. 
This can require a large batch size during training to ensure that there are a sufficient number of useful negatives~\cite{chen2020simple} -- which necessitates large compute resources during training. 
Recent advances include memory banks to address the need for large batches~\cite{wu2018unsupervised,he2020momentum,chen2020mocov2}, additional embedding layers~\cite{chen2020big}, and more advanced augmentations~\cite{swav_2020}, among others. 

In our experiments, we compare the performance of several leading self-supervised learning algorithms~\cite{chen2020simple,chen2020mocov2,swav_2020,chen2020big}  to conventional supervised learning in the context of fine-grained pretraining to try to understand what gap, if any, exists between the features learned by these very different paradigms on natural world image classification tasks.

\subsection{Benchmarking Representation Learning} 
Like Cui~\etal{}\cite{cui2018large}, we are also interested in understanding how well models trained on large-scale natural world datasets can transfer to downstream fine-grained tasks. 
However, \cite{cui2018large} only explored transfer learning using fully supervised, as opposed to self-supervised, training. 
\cite{su2019does} combined self-supervised and meta learning and showed improved few-shot classification accuracy for fine-grained categories. 
Instead of jointly training our models, we decouple feature learning from classification so that we can better understand generalization performance. 

Our work can be seen as a continuation of recent attempts to benchmark the performance of self-supervised learning \eg \cite{goyal2019scaling,kolesnikov2019revisiting,zhai2019large}.
We swap out their pretext tasks for more recent approaches and utilize natural world evaluation datasets containing a mix of fine and coarse-grained visual concepts to test the generalization of the learned features. 
This is in contrast to standard computer vision datasets or synthetic tasks~\cite{newell2020useful} that are commonly used for evaluation.  

The majority of existing self-supervised methods train on  ImageNet~\cite{russakovsky2015imagenet}. 
There are some exceptions, such as \cite{goyal2019scaling} and \cite{grill2020bootstrap} that also train on alternative datasets such as  YFCC100M~\cite{thomee2016yfcc100m} and Places205~\cite{zhou2017places}, respectively. 
We present results obtained by learning representations obtained through self-supervision alone on a large-scale natural world dataset -- as opposed to just  linear evaluation~\cite{misra2020self,swav_2020,ericsson2020well} or finetuning in this domain~\cite{he2020momentum}.

\subsection{Fine-Grained Datasets}
The vision community is not lacking in image datasets. 
The set of existing datasets include those that are large-scale and span broad category groups \eg \cite{russakovsky2015imagenet,OpenImages2}, through to smaller, but densely annotated, ones \eg \cite{everingham2010pascal,lin2014microsoft,krishna2017visual,gupta2019lvis}. 
In addition, there are a number of domain specific (\ie ``fine-grained'') datasets covering object categories such as airplanes~\cite{maji2013fine,vedaldi2014understanding}, birds~\cite{wah2011caltech,berg2014birdsnap,van2015building,krause2016unreasonable}, dogs~\cite{KhoslaYaoJayadevaprakashFeiFei_FGVC2011,parkhi12a,liu2012dog}, fashion~\cite{jia2020fashionpedia}, flowers~\cite{nilsback2006visual,nilsback2008automated},  food~\cite{bossard14,hou2017vegfru}, leaves~\cite{kumar2012leafsnap}, vehicles~\cite{krause20133d,lin2014jointly,yang2015large,gebru2017fine}, and, of course, human faces~\cite{LFWTech,parkhi2015deep,guo2016ms,cao2017vggface2}. 
Most closely related to our work are the existing iNaturalist species classification datasets~\cite{gvanhorn2018inat,iNatCompGithub}, which contain a set of coarse and fine-grained species classification problems. 

Distinct from these existing datasets, our new NeWT dataset presents a rich set of evaluation tasks that are not solely focused on one type of visual challenge \eg species classification. 
Instead, NeWT contains a wide variety of tasks encompassing behavior, health, context, among others. 
Most importantly, our tasks are informed by natural world domain experts and are thus grounded in real-world use cases. 
Paired with our new iNat2021 dataset, which contains five times more training images and nearly 20\% more categories than the largest previous version~\cite{gvanhorn2018inat}, they serve as a valuable tool to enable us to better understand and evaluate progress in both transfer and self-supervised learning in challenging visual domains.

\section{The iNaturalist 2021 Dataset}

\subsection{Dataset Overview}

\begin{table*}[t]
\small
\centering
\begin{tabular}{|l | r r r r r r r |} 
 \hline
 dataset & \# classes & \# train  & \# val & \# test & min \# ims & max \# ims & avg \# ims\\ 
 \hline\hline
 iNat2017~\cite{gvanhorn2018inat} & 5,089 & 579,184 & 95,986 & 182,707 & 9 & 3919 & 114\\ 
 iNat2018~\cite{iNatCompGithub}  & 8,142 & 437,513 & 24,426 & 149,394 & 2 & 1,000 & 54 \\
 iNat2019~\cite{iNatCompGithub}  & 1,010 & 265,213 & 3,030 & 35,350 & 16 & 500 & 263\\ \hline
 {\bf iNat2021 mini}  & 10,000 & 500,000 & $^*$100,000 & $^*$500,000 & 50 & 50 & 50\\
 {\bf iNat2021}  & 10,000 & 2,686,843 & $^*$100,000 & $^*$500,000 & 152 & 300 & 267 \\
 \hline
\end{tabular}
\vspace{5pt}
\caption{Comparison of iNat2021 dataset to previous iterations. 
iNat2021 is more than five times larger than existing large-scale species classification datasets, making it a valuable tool for benchmarking representation learning. 
Min, max, and avg refer to the number of images per class in the respective training sets. $^*$Both variants of iNat2021 use the same validation and test sets. 
}
 \vspace{-8pt}
\label{table:inat_datasets}
\end{table*}

While several large-scale natural world datasets already exist, the current largest one, iNat2017~\cite{gvanhorn2018inat}, only contains half the number of training images as ImageNet~\cite{russakovsky2015imagenet}. 
To better facilitate research in representation learning for this domain, we introduce a new  image dataset called iNat2021.  
iNat2021 consists of 2.7M training images, 100k validation images, and 500k test images, and represents images from 10k species spanning the entire tree of life. 
In addition to its overall scale, the main distinguishing feature of iNat2021 is that it contains at least 152 images in the training set for each species. 
We provide a comparison to existing datasets in Table~\ref{table:inat_datasets} and a breakdown of the image distribution in Table~\ref{tab:inat_iconic_overview}. 
Unlike previous iterations, we have split the training and testing images in iNat2021 by a specific date and have allowed a particular photographer to have images in both the train and test splits. 
There is an intuitive interpretation to this decision: we are retroactively building a computer vision training dataset, composed of data that was submitted \textit{over a year ago}, to classify the most observed species in the \textit{last year}, which is our test set. While there are many ways we could have decided the train and test split criteria, we believe this is particularly natural and lends itself well to future updates (the date split simply increases by a year). 
A detailed description of the steps we took to create the dataset are outlined in the supplementary material. 

In addition to the full sized dataset, we have also created a smaller version (iNat2021 mini) that contains 50 training images per species, sampled from the full train split. 
These two different training splits allows researchers to explore the benefits of training algorithms on five times more data. 
The mini dataset also keeps the training set size reasonable for desktop-scale experiments. 
In addition to the images themselves, we also include latitude, longitude, and time data for each, facilitating research that incorporates additional meta data to improve fine-grained classification accuracy, \eg \cite{mac2019presence,chu2019geo}.

\subsection{Comparisons to iNat2017-2019}

In Table~\ref{table:inat_datasets} we compare the new iNat2021 dataset with previous datasets built from iNaturalist. 
iNat2017 was the first large-scale species classification dataset~\cite{gvanhorn2018inat}. 
iNat2018 addressed the long tail problem inherent in large-scale media repositories. 
iNat2019 attempted to focus specifically on genera with large number of species (at least 10), resulting in a smaller dataset consisting of many $10$-way fine-grained classification problems. %$\geq10$-way fine-grained classification problems. 
Our iNat2021 dataset is similar to iNat2017 and iNat2018 in terms of its large-scale scope, however we incorporate the iNat2019 style focus on fine-grained challenges with our introduction of the NeWT collection of evaluation datasets, see Section~\ref{sec:newt}. While we have effectively removed the long tail training distribution that was the focus of other iNat datasets, we have included sufficient images per species where this phenomena can still be studied by systematically removing data. 
More data per species has the effect of decreasing the difficulty of iNat2021 in the purely supervised setting, but we believe that the additional images for each category are essential to enable us to systematically evaluate the effectiveness of self-supervised learning for natural world visual categories.

\subsection{Baseline Supervised Experiments}
\label{sec:inat_baselines}

\begin{table}[t]
\small
\centering
\begin{tabular}{|l | r r r r r |} 
 \hline
 train split & top-1 &  top-2  & top-3 & top-4 & top-5 \\ 
 \hline\hline
 iNat2021 mini  & 0.654 & 0.759 & 0.806 & 0.833 & 0.851 \\
 iNat2021       & 0.760 & 0.848 & 0.882 & 0.901 & 0.914  \\
 \hline
  iNat2021 mini * & 0.616 & 0.722 & 0.769 & 0.798 & 0.818 \\
 iNat2021 *       & 0.746 & 0.836 & 0.872 & 0.891 & 0.904  \\
 \hline
\end{tabular}
\vspace{5pt}
\caption{Top-K Accuracy on the iNat2021 test set. Models marked with a * have been initialized with random weights, otherwise ImageNet initialization is used.}
\label{table:inat_topk}
\end{table}

\begin{table}[t]
\small
% \centering
% \renewcommand{\arraystretch}{0.9}
\centering
\begin{tabular}{ |c|l|r|r|r|r| }\hline 
 & Iconic Group      &\vtop{\hbox{\strut Species}\hbox{\strut Count}}     & \vtop{\hbox{\strut Train}\hbox{\strut Images}}    &  \vtop{\hbox{\strut Full}\hbox{\strut ACC}} & \vtop{\hbox{\strut Mini }\hbox{\strut ACC}}\\\hline
\includegraphics[height=8px]{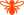}        & Insects           & 2,526 & 663,682   & 0.813 & 0.715\\
\includegraphics[height=8px]{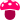}          & Fungi             & 341   & 90,048    & 0.786 & 0.707\\
\includegraphics[height=8px]{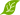}        & Plants            & 4,271 & 1,148,702 & 0.800 & 0.692\\
\includegraphics[height=8px]{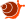}       & Mollusks          & 169   & 44,670    & 0.756 & 0.670\\
\includegraphics[height=8px]{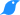}       & Animalia          & 142   & 37,042    & 0.747 & 0.654\\
\includegraphics[height=8px]{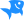} & Fish              & 183   & 45,166    & 0.725 & 0.640\\
\includegraphics[height=8px]{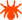}      & Arachnids         & 153   & 40,687    & 0.704 & 0.582\\
\includegraphics[height=8px]{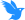}           & Birds             & 1,486 & 414,847   & 0.662 & 0.537\\
\includegraphics[height=8px]{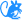}       & Mammals           & 246   & 68,917    & 0.590 & 0.496\\
\includegraphics[height=8px]{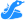}       & Reptiles          & 313   & 86,830    & 0.554 & 0.430\\
\includegraphics[height=8px]{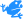}       & Amphibians        & 170   & 46,252    & 0.526 & 0.417\\
\hline %& \textbf{Total}&5,089&579,184&95,986&561,767\\\hline
\end{tabular}
\vspace{5pt}
\caption{
Number of species, training images, and mean test accuracy in iNat2021 for each iconic group. 
`Animalia' is a catch-all category that contains species that do not fit in the other iconic groups.
For the mini train split, each species has 50 train images.
}
\label{tab:inat_iconic_overview}
\end{table}

We train ResNet50~\cite{he2016deep} networks, both with and without ImageNet initialization, to benchmark the performance of iNat2021. 
%To benchmark the iNat2021 dataset we train a ResNet50~\cite{he2016deep} model using the pytorch ImageNet example training script~\footnote{https://github.com/pytorch/examples/blob/master/imagenet/main.py}. 
Table~\ref{table:inat_topk} shows the top-k accuracy achieved when training using the full and mini datasets, and Table~\ref{tab:inat_iconic_overview} shows the top-1 accuracy broken down by iconic groups. 
The model trained on the mini dataset results in a top-1 accuracy of 65.4\%, while the full model achieves 76.0\%, showing that an increase from 500k training images to 2.7M results in an $\sim$11 percentage point increase in accuracy. 
The corresponding top-1 results for the validation set are 65.8\% and 76.4\%. 
On average, insects are the best performing iconic group, and amphibians are the worst performing group. While these average statistics are interesting, we do not believe they demonstrate that insects are necessarily ``easier'' to identify than amphibians. We are most likely seeing a bias in the iNat2021 dataset. Perhaps, on average, it is easier to take a close-up photograph of an insect than it is to photograph an amphibian. Or perhaps the amphibian species have more visual modalities than insects. % (\eg eggs, tadpoles, adults, road kill, etc.). 
Finally, we observe that models trained from randomly initialized weights perform slightly worse than those trained from ImageNet initialization, but the gap closes when training on the full dataset. 

\section{NeWT: Natural World Tasks}
\label{sec:newt}
\begin{figure*}[t]
\vspace{-10pt}
\centering
\includegraphics[trim=50 0 60 0,clip,width=\textwidth]{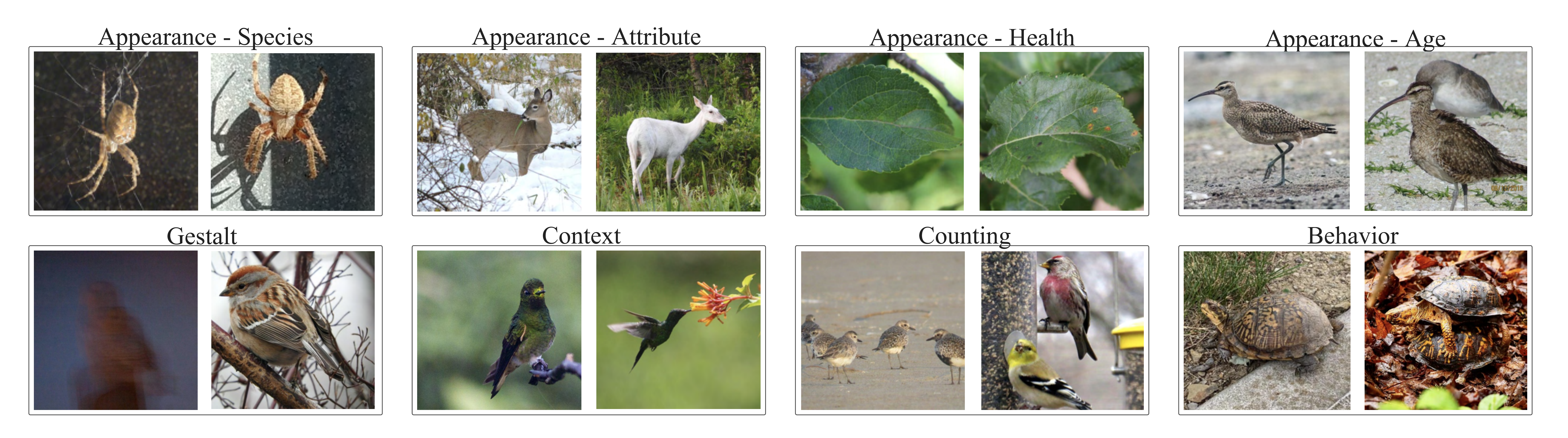}
\vspace{-20pt}
 \caption{Example image pairs from a binary classification task within each coarse task grouping of the NeWT dataset
 }
 \vspace{-5pt}
\label{fig:eval_tasks}
\end{figure*}

Large media repositories, such as Flickr, the Macaulay Library, and iNaturalist have been utilized to create species classification datasets such as CUB~\cite{wah2011caltech}, BirdSnap~\cite{berg2014birdsnap}, NABirds~\cite{van2015building}, and the collection of iNaturalist competition datasets~\cite{gvanhorn2018inat}. 
These datasets have become standard experimental resources for computer vision researchers and have been used to benchmark the progress of classification models over the last decade. Improvements on these datasets have in turn led to the incorporation of these models into useful applications that assist everyday users in recognizing the wildlife around them, \eg~\cite{kumar2012leafsnap,seek_2020,merlin_2020}. 
However, there are far more questions that biologists and practitioners would like to ask of these large media repositories in addition to ``What species is in this photo?'' 
For example, an ornithologist may like to ask, ``Does this photo contain a nest?'' or ``Does this photo show an adult feeding a nestling?'' Similarly, a herpetologist may like to ask, ``Does this photo show mating behavior for the Southern Alligator Lizard?'' 
Researchers can certainly answer these questions themselves for a few images. 
The problem is the scale of these archives, and the fact that they are continually growing. Can a computer vision model be used to answer these questions? While we do not have large collections of datasets labeled with nests or eggs or mating behavior, we do have large-scale species classification datasets. This raises the question about the adaptability of a model trained for species classification to these new types of questions. Similarly, with the recent advances in self-supervised learning there is the potential for a self-supervised model to be readily adapted to answer these varied tasks. 
To help address these questions we have constructed a collection of Natural World Tasks (NeWT) that can be used to benchmark current representation learning methods.

NeWT is comprised of 164 highly curated binary classification tasks sourced from iNaturalist, the Macaulay Library, and the NABirds dataset, among others. 
No images from NeWT occur in the iNat2021 training dataset, and the images in tasks not sourced from iNaturalist are reasonably similar to images found on iNaturalist. 
This makes the iNat2021 dataset a perfect pretraining dataset for NeWT. 
Unlike some of the potential data quality issues found in iNat2021 (see supplementary material), each task in NeWT has been vetted for data quality with the assistance of domain experts. While species classification still plays a large role in NeWT (albeit reduced down to difficult fine-grained pairs of species), the addition of other types of tasks makes this dataset uniquely positioned to determine how well different pretrained models can answer various natural world questions.  Each task has approximately uniform positive and negative samples, as well as approximately uniform train and test samples. The size of each task is modest, on the order of 50-100 images per class per split (for a total of 200-400 images per task), which makes them very convenient for training and evaluating linear classifiers. We have coarsely categorized the tasks into eight groups (see Figure~\ref{fig:eval_tasks} for visual examples) with the total number of binary tasks per group in parentheses:
\vspace{-5pt}
\begin{itemize}[leftmargin=*]
    \itemsep -4pt 
    \item \textbf{Appearance - Age (14)} Tasks where the age of the species is the decision criteria, \eg  ``Is this a hatch-year Whimbrel?'' 
    \item \textbf{Appearance - Attribute (7)}: Tasks where a specific attribute of an organism is used to make the decision, \eg  ``Is the deer leucistic?'' 
    \item \textbf{Appearance - Health (9)}: Tasks where the health of the organism is the decision criteria, \eg  ``Is the plant diseased?'' 
    \item \textbf{Appearance - Species (102)}: Tasks where the goal is to distinguish two visually similar species. This can include species from iNat2021, but with new, unseen training data, and tasks from species not included in iNat2021.
    \item \textbf{Behavior (16)} Tasks where the evidence of a behavior is the decision criteria, \eg  ``Are the lizards mating?''  
    \item \textbf{Context (8)} Tasks where the immediate or surrounding context of the organism is the decision criteria, \eg  ``Is the hummingbird feeding at a flower?'' 
    \item \textbf{Counting (2)} Tasks where the number of specific instances is the decision criteria, \eg  ``Are there multiple bird species present?''  
    \item \textbf{Gestalt (6)} Tasks where the quality, composition, or type of photo is the decision criteria, \eg  ``Is this a high quality or low quality photograph of a bird?''  
\end{itemize}

\section{Experiments} 
\label{sec:experiments}
\begin{figure*}[t]
\centering
\includegraphics[width=\linewidth]{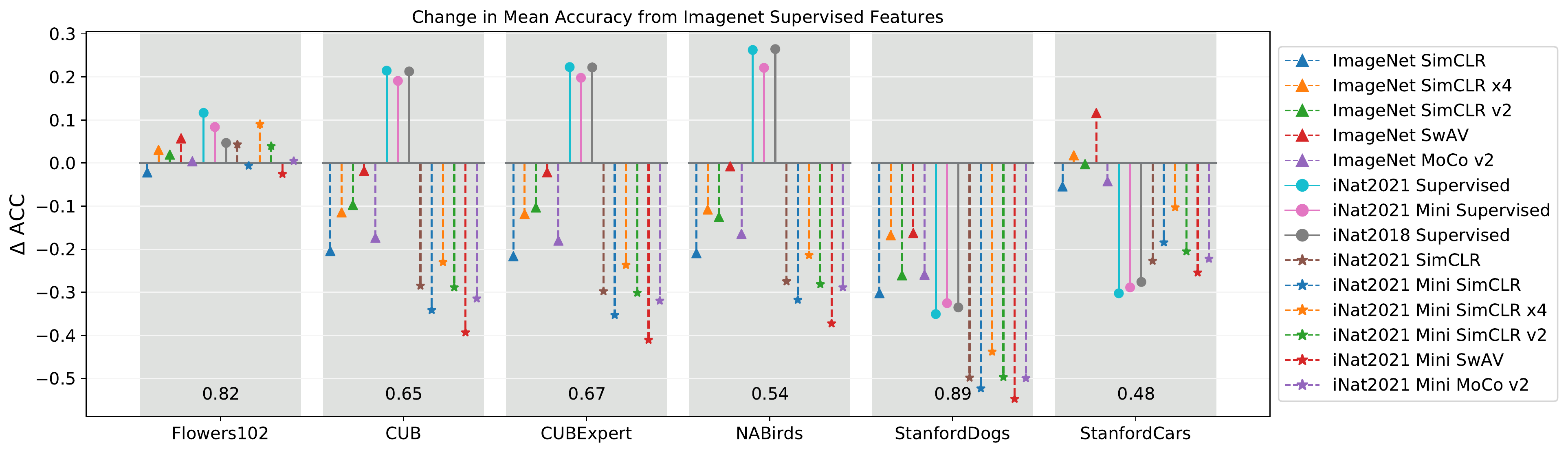}
\vspace{-15pt}
\caption{{\bf Fine-grained evaluation.} The mean top-1 accuracy difference between ``off-the-shelf'' supervised ImageNet features and other pretraining strategies on existing fine-grained datasets. For context, the accuracy of the ImageNet features are printed above the dataset labels along the x-axis. All methods utilize a ResNet50 backbone architecture, and all experiments use features extracted by the last convolution block (dim=2048) to train a linear SVM using SGD (x4 models have dim=8192). 
%We use 3-fold cross-validation to choose an appropriate regularization parameter per dataset.
Techniques that make use of supervised pretraining have a solid stem line, while techniques that use self-supervision for pretraining have a dashed stem line. Techniques that utilize ImageNet have a triangle marker, techniques that utilize an iNat dataset with supervision have a circle marker, and techniques that utilize an iNat dataset with a self-supervision training objective have a star marker.  
Several patterns are apparent: (1) Self-supervised methods rarely do better than ``off-the-shelf'' supervised ImageNet features. (2) Pretraining on iNat datasets with supervision leads to better results on downstream tasks that contain categories similar to those found in iNat datasets (\ie~flowers and birds), but this does not hold for self-supervised objectives. (3) Self-supervised models trained on ImageNet do better than their iNat counterparts. For detailed accuracy numbers see the supplementary material.}
\label{fig:fg_diff_from_imagenet}
\end{figure*}

Here we present an analysis of different learned image representations trained on multiple datasets and evaluate their effectiveness on existing fine-grained datasets and NeWT. 

\subsection{Implementation Details}
\vspace{-5pt}
Given a specific configuration of \{feature extractor, pretraining dataset, training objective\}, our feature representation evaluation protocol is the same for all experiments. Every experiment uses the ResNet50~\cite{he2016deep} model as the feature extractor, with some experiments modifying the width multiplier parameter of the network to $4$.  We consider ImageNet, iNat2018, iNat2021, and the iNat2021 mini dataset for the pretraining dataset. The training objective can either be a supervised classification loss (standard cross-entropy) or one of the following self-supervised objectives: SimCLR~\cite{chen2020simple}, SimCLR v2~\cite{chen2020big}, SwAV~\cite{swav_2020}, or MoCo v2~\cite{chen2020mocov2}. 

The supervised experiments using iNat2021 mini and iNat2018 are trained for 65-90 epochs, starting from ImageNet initialization, and we used the model checkpoint that performed the best on the respective validation set. The supervised experiments using iNat2021 were trained for 20 epochs, also starting from ImageNet initialization. For self-supervised techniques pretrained on ImageNet, we make use of model checkpoint files accompanying the official implementation of the method. For models self-supervised on iNat datasets we used default parameters from the respective techniques unless otherwise stated. Our experiments using SimCLR v2 on iNat datasets do not incorporate knowledge distillation from a larger network nor the MoCo style memory mechanism; instead we train the ResNet50 backbone using a 3-layer projection head instead of the 2-layer projection head found in the original SimCLR objective. See the supplementary material for additional details on model training.

\begin{figure*}[ht]
\centering
\includegraphics[width=\linewidth]{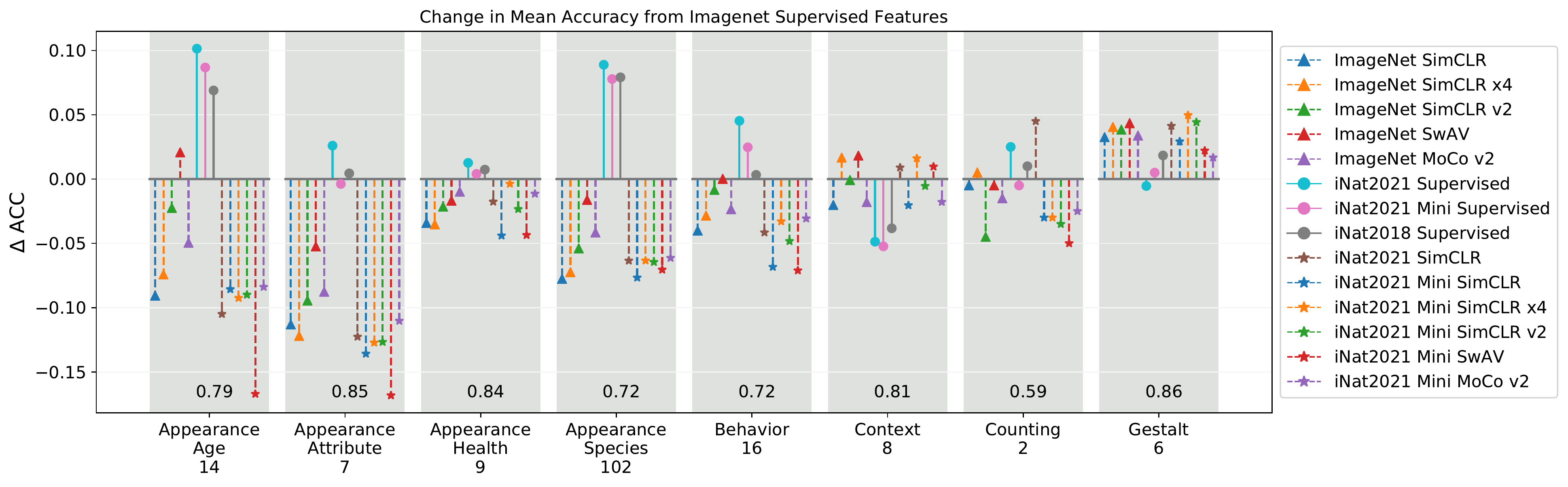}
\vspace{-15pt}
\caption{{\bf NeWT evaluation.} The mean top-1 accuracy difference between ``off-the-shelf'' supervised ImageNet features and various other pretraining strategies on the NeWT dataset, divided into related groups. See Figure~\ref{fig:fg_diff_from_imagenet} for information regarding the plot organization and interpretation. 
Several patterns are apparent: (1) Supervised learning using iNaturalist data achieves better performance on NeWT tasks that focus on species appearance and behavior. (2) Self-supervised learning achieves better performance compared to supervised methods on the \emph{Gestalt} tasks, \ie~tasks that do not focus on a particular individual. (3) For self-supervision, we do not see a consistent benefit to using iNat2021 over ImageNet (unlike Figure~\ref{fig:fg_diff_from_imagenet}); sometimes pretraining on iNat2021 leads to better performance than pretraining on ImageNet, other times it is reversed.
For detailed accuracy numbers see the supplementary material.
}
\label{fig:newt_diff_from_imagenet}
\end{figure*}

After training the ResNet50 model on the selected dataset, it is then used as a feature extractor on ``downstream'' evaluation datasets. Images are resized so the smaller edge is 256 then we take a center crop of 224x224, which is then passed through the model. No other form of augmentation is used. Features are extracted from the last convolutional block of the ResNet50 model and have a dimension of $2048$ unless the width of the network was modified to $4$, in which case the dimension is $8192$. A linear model is then trained on these features and the associated ground truth class labels. 
Details of the linear model are provided below. 
We use top-1 accuracy on the held out test set of the respective ``downstream'' dataset as the evaluation metric for the linear model. We compare different feature representations by measuring the relative change in accuracy when using supervised ImageNet features as the baseline ($\Delta$ ACC in Figure~\ref{fig:fg_diff_from_imagenet} and Figure~\ref{fig:newt_diff_from_imagenet}). We chose supervised ImageNet features as the baseline because these features are readily accessible to nearly all practitioners, requiring zero additional training and very little computational resources. To facilitate reproducibility, all pretrained models are accessible from our GitHub project page.

\subsection{Experiments on Fine-Grained Datasets}
\vspace{-5pt}
In this section we demonstrate the utility of iNat2021 as a pretraining dataset for existing fine-grained datasets. The extracted features are evaluated on Flowers102~\cite{nilsback2008automated}, CUB~\cite{wah2011caltech}, NABirds~\cite{van2015building}, StanfordDogs~\cite{KhoslaYaoJayadevaprakashFeiFei_FGVC2011}, and StanfordCars~\cite{krause20133d}. We also present results on CUBExpert, which is the standard CUB dataset but the class labels have been verified and cleaned by domain experts~\cite{van2015building}. For these experiments, the linear model is a SVM trained using SGD for a maximum of $3\text{k}$ epochs with a stopping criteria tolerance of $1\mathrm{e}{-5}$. 
For every experiment, we use 3-fold cross validation to determine the appropriate regularization constant $\alpha \in [1\mathrm{e}{-6}, 1\mathrm{e}{-5}, 1\mathrm{e}{-4}, 1\mathrm{e}{-3}, 1\mathrm{e}{-2}, 0.1, 1, 10]$.

We present the relative accuracy changes in relation to supervised ImageNet features for the various techniques in Figure~\ref{fig:fg_diff_from_imagenet}. 
Please consult the supplementary material for specific accuracy values. 
Overall we find that supervised techniques produce the best features for all datasets except Stanford Cars, where the SwAV model trained on ImageNet produced the best features. The iNat2021 supervised model is the best performing on Flowers102, CUB, and CUBExpert; the iNat2018 supervised model is the best on NABirds, narrowly eclipsing the iNat2021 supervised model ($0.806$ vs. $0.804$ top-1 accuracy); and the supervised ImageNet model is the best on StanfordDogs. 
When considering self-supervised methods, the SwAV model trained on ImageNet is consistently the top performer except for the Flowers102 dataset, where the SimCLR x4 model trained on iNat2021 mini achieves better performance (using a 4x larger feature vector than the SwAV model). 

In terms of pretraining datasets for self-supervised techniques, the ImageNet dataset appears better than the iNat2021 dataset: note the lines for self-supervised methods trained on iNat2021 and iNat2021 mini in Figure~\ref{fig:fg_diff_from_imagenet} are uniformly below their ImageNet counterparts for all datasets except Flowers102. While not particular surprising for the Stanford Dogs and Cars datasets that differ fundamentally from the iNaturalist domain, this is a surprising result for the bird datasets: CUB, CUBExpert, and NABirds.  The ImageNet dataset has about 60 species of birds with $\sim$60k training images, while the iNat2021 dataset has 1,486 species with 414,847 and 74,300 training images in the large and mini splits respectively. Even with increased species and training samples, the ImageNet dataset out performs the iNat2021 dataset on downstream bird tasks. Perhaps this is an artifact of the \textit{types} of images within these datasets as opposed to the \textit{domain} of the datasets. The self-supervised techniques considered in this work were designed for ImageNet, therefore their default augmentation strategy appears to be designed for objects that take up a large fraction of the image size. Applying these strategies to datasets where objects do not necessarily take up large fraction of the image size (like iNat2021) appears to be inappropriate. See the supplementary material for an analysis of the sizes of bird bounding boxes across the datasets. 

Note that supervised methods can still recover discriminative features from the iNat datasets (see the performance of supervised iNat2021 and iNat2021 mini in Figure~\ref{fig:fg_diff_from_imagenet}), so it should be feasible for self-supervised methods to leverage these datasets to learn better representations. Interestingly, the effect of data size is not very apparent in Figure~\ref{fig:fg_diff_from_imagenet} for the experiments that use the large and mini variants of the iNat2021 dataset. While performance on the actual iNat2021 improved by 11 percentage points when switching from the mini to the large (see Table~\ref{table:inat_topk}), we do not see a similar level of improvement for downstream tasks.

\subsection{Experiments on NeWT}
\vspace{-5pt}
In this section we use the collection of binary tasks in NeWT as ``downstream'' classification tasks to investigate the effect of different pretraining methods. For these experiments the linear model is a SVM trained using liblinear for a maximum of $1\text{k}$ iterations with a stopping criteria tolerance of $1\mathrm{e}{-5}$. For every experiment, we use 3-fold cross validation to determine the appropriate regularization constant $\text{C} \in [1\mathrm{e}{-4}, 1\mathrm{e}{-3}, 1\mathrm{e}{-2}, 0.1, 1, 10, 1\mathrm{e}{2},1\mathrm{e}{3}]$.

The supervised ImageNet model achieved an average accuracy of $0.744$ across all 164 NeWT tasks. The supervised iNat2021 model achieved the best average accuracy with a score of $0.806$, followed by the supervised iNat2021 mini model at $0.793$ and then the supervised iNat2018 model at $0.791$. For self-supervised models, the SwAV model trained on ImageNet did the best at $0.733$ average accuracy. We show the relative accuracy changes in relation to supervised ImageNet features for the various techniques in Figure~\ref{fig:newt_diff_from_imagenet}, see the supplementary material for specific accuracy values.

For the \emph{Appearance} based tasks in NeWT (which focus on a specific individual in the photo), we can see that there is a clear benefit to doing supervised pretraining on data from iNaturalist (using either iNat2018, iNat2021, or iNat2021 mini). 
\emph{Species} classification, unsurprisingly, and \emph{Age} have the biggest improvement followed by \emph{Attribute} and then \emph{Health}. We do not see the same benefit when using self-supervision for these \emph{Appearance} based tasks. We instead find self-supervised models performing worse on average than ImageNet supervised features, even though they are trained on data from iNaturalist. 
Similarly, the \emph{Behavior} tasks benefited from supervised pretraining on iNat datasets, but did not benefit from self-supervised pretraining. 
No method significantly improved performance on the \emph{Context} tasks compared to supervised ImageNet features. 
All methods did relatively poorly on the two \emph{Counting} tasks ($0.59$ baseline performance, note that chance is 50\%). This could highlight the inappropriateness of using a classifier for detection style tasks, or it could highlight a particularly disappointing generalization behavior of these models. The SimCLR method trained on iNat2021 is a notable outlier in this experiment but the reason is unclear. 
Interestingly, all self-supervised models appear to provide a benefit over supervised ImageNet features and supervised iNat features for the \emph{Gestalt} tasks, where the whole image needs to be analyzed as opposed to focusing on a particular subject.

Similar to the fine-grained datasets result, we see a reduced improvement between the iNat2021 large and mini datasets on the NeWT tasks as compared to evaluating on the iNat2021 test set. The SimCLR model achieved $0.678$ mean accuracy using the iNat2021 mini split, and $0.689$ with the full dataset. The supervised model went from $0.793$ mean accuracy to $0.806$. This result is surprising given the typical expectation of performance improvement when training with more data. Goyal \etal~\cite{goyal2019scaling} perform experiments where they scale the amount of training data by a factor of 10, 50, and 100 and they see a larger performance gain for the ResNet50 model, albeit using Jigsaw~\cite{noroozi2016unsupervised} and Colorization~\cite{zhang2016colorful} as pretext tasks, and Pascal VOC07~\cite{everingham2010pascal} as the downstream task. So either 5x more data is not a sufficient data increase, or self-supervision objectives like SimCLR behave differently.

While the experiments on existing fine-grained datasets in Figure~\ref{fig:fg_diff_from_imagenet} showed a benefit to using ImageNet over iNat2021 as the pretraining dataset for self-supervision, the NeWT results are much more mixed. For example SimCLR trained using ImageNet achieves better performance on average for the \emph{Appearance - Age} tasks than SimCLR trained using iNat2021 ($0.702$ vs $0.688$), but the results are flipped for the \emph{Appearance - Species} tasks ($0.647$ vs $0.661$).

\subsection{Discussion}
\vspace{-5pt}
We summarize our main findings:

\noindent \textbf{Supervised ImageNet features are a strong baseline.} The off-the-shelf supervised ImageNet features were often much better than the features derived from self-supervised models trained on either ImageNet or iNat2021. This applies to supervised iNat2021 features as well. It is currently easier to achieve downstream performance gains from a model trained with a supervised objective (assuming it is possible to get labels).

\noindent \textbf{Fine-grained classification is challenging for self-supervised models.} For most self-supervised methods performance is not close to supervised methods for the fine-grained datasets tested, see Figure~\ref{fig:fg_diff_from_imagenet}. However, the SwAV method has closed the gap and is better in some cases (\eg~Stanford Cars). This trend did not hold when SwAV was trained on iNat2021 mini data.

\noindent \textbf{Not all tasks are equal.} Self-supervised features can be more effective compared to supervised ones for certain tasks (\eg~see the \emph{Gestalt} tasks in NeWT in Figure~\ref{fig:newt_diff_from_imagenet}). This highlights the value of benchmarking performance on a varied set of classification tasks, in addition to conventional object classification. 

\noindent \textbf{More data does not help methods as much for downstream tasks.} While we observe a large boost in accuracy on the iNat2021 test set when we increase the amount of training data (+11 percentage points, see Tables~\ref{table:inat_topk} and \ref{tab:inat_iconic_overview}), this boost is much smaller for both supervised and self-supervised models on the fine-grained datasets and NeWT (see the differences between iNat2021 large and mini for the supervised and SimCLR experiments in Figures~\ref{fig:fg_diff_from_imagenet} and~\ref{fig:newt_diff_from_imagenet}).

\noindent \textbf{Self-supervised ImageNet training settings do not necessarily generalize.} The performance gap between supervised and self-supervised features on downstream tasks is closing when the feature extractor is trained on ImageNet. However, the gap between supervised and self-supervised features is much larger when the the feature extractor is trained on iNat2021. This potentially points to self-supervised training settings being overfit to ImageNet \eg via hyperparameters or the image augmentations used.

\section{Conclusion}
\vspace{-8pt}
We presented, and benchmarked, the iNat2021 and NeWT datasets. The iNat2021 dataset contains 2.7M training images covering 10k species. As a large-scale image dataset we have shown its utility as a powerful pretraining network for a variety of existing fine-grained datasets  as well as the NeWT dataset. Our NeWT dataset expands beyond the question of ``What species is this?'', to incorporate questions that challenge models to identify behaviors, health, and context questions as they relate to wildlife captured in photographs. 
Our experiments on NeWT reveal interesting performance differences between supervised and self-supervised learning methods. 
While supervised learning appears to still have an edge over existing self-supervised approaches, new methods are constantly being introduced by the research community. The iNat2021 and NeWT datasets should serve as a valuable resource for benchmarking these new techniques as they expose challenges not present in the standard  datasets currently in use.

\vspace{5pt}
\noindent{\bf Acknowledgments}
Thanks to the iNaturalist team and community for providing access to data, Eliot Miller and Mitch Barry for helping to curate NeWT, and to Pietro Perona for valuable feedback.

{\small
\bibliographystyle{ieee_fullname}
\bibliography{main}

\begin{thebibliography}{10}\itemsep=-1pt

\bibitem{iNatWeb}
{iNaturalist}.
\newblock \url{www.inaturalist.org}, accessed Nov 14 2020.

\bibitem{iNatCompGithub}
{iNaturalist Challenge Datasets}.
\newblock \url{https://github.com/visipedia/inat_comp}, accessed Nov 14 2020.

\bibitem{berg2014birdsnap}
Thomas Berg, Jiongxin Liu, Seung~Woo Lee, Michelle~L Alexander, David~W Jacobs,
  and Peter~N Belhumeur.
\newblock Birdsnap: Large-scale fine-grained visual categorization of birds.
\newblock In {\em CVPR}, 2014.

\bibitem{bossard14}
Lukas Bossard, Matthieu Guillaumin, and Luc Van~Gool.
\newblock Food-101 -- mining discriminative components with random forests.
\newblock In {\em ECCV}, 2014.

\bibitem{cao2017vggface2}
Q. Cao, L. Shen, W. Xie, O.~M. Parkhi, and A. Zisserman.
\newblock Vggface2: A dataset for recognising faces across pose and age.
\newblock In {\em International Conference on Automatic Face and Gesture
  Recognition}, 2018.

\bibitem{cardinale2012biodiversity}
Bradley~J Cardinale, J~Emmett Duffy, Andrew Gonzalez, David~U Hooper, Charles
  Perrings, Patrick Venail, Anita Narwani, Georgina~M Mace, David Tilman,
  David~A Wardle, et~al.
\newblock Biodiversity loss and its impact on humanity.
\newblock {\em Nature}, 2012.

\bibitem{swav_2020}
Mathilde Caron, Ishan Misra, Julien Mairal, Priya Goyal, Piotr Bojanowski, and
  Armand Joulin.
\newblock Unsupervised learning of visual features by contrasting cluster
  assignments.
\newblock In {\em NeurIPS}, 2020.

\bibitem{chen2020simple}
Ting Chen, Simon Kornblith, Mohammad Norouzi, and Geoffrey Hinton.
\newblock A simple framework for contrastive learning of visual
  representations.
\newblock In {\em ICML}, 2020.

\bibitem{chen2020big}
Ting Chen, Simon Kornblith, Kevin Swersky, Mohammad Norouzi, and Geoffrey
  Hinton.
\newblock Big self-supervised models are strong semi-supervised learners.
\newblock In {\em NeurIPS}, 2020.

\bibitem{chen2020mocov2}
Xinlei Chen, Haoqi Fan, Ross Girshick, and Kaiming He.
\newblock Improved baselines with momentum contrastive learning.
\newblock {\em arXiv:2003.04297}, 2020.

\bibitem{chu2019geo}
Grace Chu, Brian Potetz, Weijun Wang, Andrew Howard, Yang Song, Fernando
  Brucher, Thomas Leung, and Hartwig Adam.
\newblock Geo-aware networks for fine-grained recognition.
\newblock In {\em ICCV Workshops}, 2019.

\bibitem{cui2018large}
Yin Cui, Yang Song, Chen Sun, Andrew Howard, and Serge Belongie.
\newblock Large scale fine-grained categorization and domain-specific transfer
  learning.
\newblock In {\em CVPR}, 2018.

\bibitem{doersch2015unsupervised}
Carl Doersch, Abhinav Gupta, and Alexei~A Efros.
\newblock Unsupervised visual representation learning by context prediction.
\newblock In {\em ICCV}, 2015.

\bibitem{donahue2014decaf}
Jeff Donahue, Yangqing Jia, Oriol Vinyals, Judy Hoffman, Ning Zhang, Eric
  Tzeng, and Trevor Darrell.
\newblock Decaf: A deep convolutional activation feature for generic visual
  recognition.
\newblock In {\em ICML}, 2014.

\bibitem{ericsson2020well}
Linus Ericsson, Henry Gouk, and Timothy~M Hospedales.
\newblock How well do self-supervised models transfer?
\newblock In {\em CVPR}, 2021.

\bibitem{everingham2010pascal}
Mark Everingham, Luc Van~Gool, Christopher~KI Williams, John Winn, and Andrew
  Zisserman.
\newblock The pascal visual object classes (voc) challenge.
\newblock {\em IJCV}, 2010.

\bibitem{gebru2017fine}
Timnit Gebru, Jonathan Krause, Yilun Wang, Duyun Chen, Jia Deng, and Li
  Fei-Fei.
\newblock Fine-grained car detection for visual census estimation.
\newblock In {\em AAAI}, 2017.

\bibitem{giaimo_NYT_lizard}
Ciara Giaimo.
\newblock Hold me, squeeze me, bite my head.
\newblock {\em The New York Times}, Sep 2020.

\bibitem{gidaris2018unsupervised}
Spyros Gidaris, Praveer Singh, and Nikos Komodakis.
\newblock Unsupervised representation learning by predicting image rotations.
\newblock In {\em ICLR}, 2018.

\bibitem{goyal2019scaling}
Priya Goyal, Dhruv Mahajan, Abhinav Gupta, and Ishan Misra.
\newblock Scaling and benchmarking self-supervised visual representation
  learning.
\newblock In {\em ICCV}, 2019.

\bibitem{grill2020bootstrap}
Jean-Bastien Grill, Florian Strub, Florent Altch{\'e}, Corentin Tallec, Pierre
  Richemond, Elena Buchatskaya, Carl Doersch, Bernardo Avila~Pires, Zhaohan
  Guo, Mohammad Gheshlaghi~Azar, et~al.
\newblock Bootstrap your own latent-a new approach to self-supervised learning.
\newblock {\em NeurIPS}, 2020.

\bibitem{guo2016ms}
Yandong Guo, Lei Zhang, Yuxiao Hu, Xiaodong He, and Jianfeng Gao.
\newblock Ms-celeb-1m: A dataset and benchmark for large-scale face
  recognition.
\newblock In {\em ECCV}, 2016.

\bibitem{gupta2019lvis}
Agrim Gupta, Piotr Dollar, and Ross Girshick.
\newblock {LVIS: A dataset for large vocabulary instance segmentation}.
\newblock In {\em CVPR}, 2019.

\bibitem{gutmann2010noise}
Michael Gutmann and Aapo Hyv{\"a}rinen.
\newblock Noise-contrastive estimation: A new estimation principle for
  unnormalized statistical models.
\newblock In {\em AISTATS}, 2010.

\bibitem{hadsell2006dimensionality}
Raia Hadsell, Sumit Chopra, and Yann LeCun.
\newblock Dimensionality reduction by learning an invariant mapping.
\newblock In {\em CVPR}, 2006.

\bibitem{he2020momentum}
Kaiming He, Haoqi Fan, Yuxin Wu, Saining Xie, and Ross Girshick.
\newblock Momentum contrast for unsupervised visual representation learning.
\newblock In {\em CVPR}, 2020.

\bibitem{he2016deep}
Kaiming He, Xiangyu Zhang, Shaoqing Ren, and Jian Sun.
\newblock Deep residual learning for image recognition.
\newblock In {\em CVPR}, 2016.

\bibitem{hou2017vegfru}
Saihui Hou, Yushan Feng, and Zilei Wang.
\newblock Vegfru: A domain-specific dataset for fine-grained visual
  categorization.
\newblock In {\em ICCV}, 2017.

\bibitem{LFWTech}
Gary~B. Huang, Manu Ramesh, Tamara Berg, and Erik Learned-Miller.
\newblock Labeled faces in the wild: A database for studying face recognition
  in unconstrained environments.
\newblock Technical report, University of Massachusetts, Amherst, 2007.

\bibitem{seek_2020}
iNaturalist.
\newblock {Seek by iNaturalist}, 2020.
\newblock \url{https://apps.apple.com/us/app/seek-by-inaturalist/id1353224144}.

\bibitem{jia2020fashionpedia}
Menglin Jia, Mengyun Shi, Mikhail Sirotenko, Yin Cui, Claire Cardie, Bharath
  Hariharan, Hartwig Adam, and Serge Belongie.
\newblock Fashionpedia: Ontology, segmentation, and an attribute localization
  dataset.
\newblock In {\em ECCV}, 2020.

\bibitem{KhoslaYaoJayadevaprakashFeiFei_FGVC2011}
Aditya Khosla, Nityananda Jayadevaprakash, Bangpeng Yao, and Li Fei-Fei.
\newblock Novel dataset for fine-grained image categorization.
\newblock In {\em First Workshop on Fine-Grained Visual Categorization}, 2011.

\bibitem{kolesnikov2019revisiting}
Alexander Kolesnikov, Xiaohua Zhai, and Lucas Beyer.
\newblock Revisiting self-supervised visual representation learning.
\newblock In {\em CVPR}, 2019.

\bibitem{kornblith2019better}
Simon Kornblith, Jonathon Shlens, and Quoc~V Le.
\newblock Do better imagenet models transfer better?
\newblock In {\em CVPR}, 2019.

\bibitem{OpenImages2}
Ivan Krasin, Tom Duerig, Neil Alldrin, Vittorio Ferrari, Sami Abu-El-Haija,
  Alina Kuznetsova, Hassan Rom, Jasper Uijlings, Stefan Popov, Shahab Kamali,
  Matteo Malloci, Jordi Pont-Tuset, Andreas Veit, Serge Belongie, Victor Gomes,
  Abhinav Gupta, Chen Sun, Gal Chechik, David Cai, Zheyun Feng, Dhyanesh
  Narayanan, and Kevin Murphy.
\newblock Openimages: A public dataset for large-scale multi-label and
  multi-class image classification.
\newblock {\em Dataset available from
  https://storage.googleapis.com/openimages/web/index.html}, 2017.

\bibitem{krause2016unreasonable}
Jonathan Krause, Benjamin Sapp, Andrew Howard, Howard Zhou, Alexander Toshev,
  Tom Duerig, James Philbin, and Li Fei-Fei.
\newblock The unreasonable effectiveness of noisy data for fine-grained
  recognition.
\newblock In {\em ECCV}, 2016.

\bibitem{krause20133d}
Jonathan Krause, Michael Stark, Jia Deng, and Li Fei-Fei.
\newblock 3d object representations for fine-grained categorization.
\newblock In {\em ICCV Workshops}, 2013.

\bibitem{krishna2017visual}
Ranjay Krishna, Yuke Zhu, Oliver Groth, Justin Johnson, Kenji Hata, Joshua
  Kravitz, Stephanie Chen, Yannis Kalantidis, Li-Jia Li, David~A Shamma, et~al.
\newblock Visual genome: Connecting language and vision using crowdsourced
  dense image annotations.
\newblock {\em IJCV}, 2017.

\bibitem{kumar2012leafsnap}
Neeraj Kumar, Peter~N Belhumeur, Arijit Biswas, David~W Jacobs, W~John Kress,
  Ida~C Lopez, and Jo{\~a}o~VB Soares.
\newblock Leafsnap: A computer vision system for automatic plant species
  identification.
\newblock In {\em ECCV}. 2012.

\bibitem{lin2014microsoft}
Tsung-Yi Lin, Michael Maire, Serge Belongie, James Hays, Pietro Perona, Deva
  Ramanan, Piotr Doll{\'a}r, and C~Lawrence Zitnick.
\newblock {Microsoft COCO: Common objects in context}.
\newblock In {\em ECCV}, 2014.

\bibitem{lin2014jointly}
Yen-Liang Lin, Vlad~I Morariu, Winston Hsu, and Larry~S Davis.
\newblock Jointly optimizing 3d model fitting and fine-grained classification.
\newblock In {\em ECCV}. 2014.

\bibitem{liu2012dog}
Jiongxin Liu, Angjoo Kanazawa, David Jacobs, and Peter Belhumeur.
\newblock Dog breed classification using part localization.
\newblock In {\em ECCV}. 2012.

\bibitem{mac2019presence}
Oisin Mac~Aodha, Elijah Cole, and Pietro Perona.
\newblock Presence-only geographical priors for fine-grained image
  classification.
\newblock In {\em ICCV}, 2019.

\bibitem{maji2013fine}
Subhransu Maji, Esa Rahtu, Juho Kannala, Matthew Blaschko, and Andrea Vedaldi.
\newblock Fine-grained visual classification of aircraft.
\newblock {\em arXiv:1306.5151}, 2013.

\bibitem{misra2020self}
Ishan Misra and Laurens van~der Maaten.
\newblock Self-supervised learning of pretext-invariant representations.
\newblock In {\em CVPR}, 2020.

\bibitem{newell2020useful}
Alejandro Newell and Jia Deng.
\newblock How useful is self-supervised pretraining for visual tasks?
\newblock In {\em CVPR}, 2020.

\bibitem{nilsback2006visual}
Maria-Elena Nilsback and Andrew Zisserman.
\newblock A visual vocabulary for flower classification.
\newblock In {\em CVPR}, 2006.

\bibitem{nilsback2008automated}
Maria-Elena Nilsback and Andrew Zisserman.
\newblock Automated flower classification over a large number of classes.
\newblock In {\em Indian Conference on Computer Vision, Graphics \& Image
  Processing}, 2008.

\bibitem{noroozi2016unsupervised}
Mehdi Noroozi and Paolo Favaro.
\newblock Unsupervised learning of visual representations by solving jigsaw
  puzzles.
\newblock In {\em ECCV}, 2016.

\bibitem{parkhi2015deep}
Omkar~M Parkhi, Andrea Vedaldi, Andrew Zisserman, et~al.
\newblock Deep face recognition.
\newblock In {\em BMVC}, 2015.

\bibitem{parkhi12a}
O.~M. Parkhi, A. Vedaldi, A. Zisserman, and C.~V. Jawahar.
\newblock Cats and dogs.
\newblock In {\em CVPR}, 2012.

\bibitem{russakovsky2015imagenet}
Olga Russakovsky, Jia Deng, Hao Su, Jonathan Krause, Sanjeev Satheesh, Sean Ma,
  Zhiheng Huang, Andrej Karpathy, Aditya Khosla, Michael Bernstein, et~al.
\newblock Imagenet large scale visual recognition challenge.
\newblock {\em IJCV}, 2015.

\bibitem{su2019does}
Jong-Chyi Su, Subhransu Maji, and Bharath Hariharan.
\newblock When does self-supervision improve few-shot learning?
\newblock In {\em ECCV}, 2020.

\bibitem{sullivan2009ebird}
Brian~L Sullivan, Christopher~L Wood, Marshall~J Iliff, Rick~E Bonney, Daniel
  Fink, and Steve Kelling.
\newblock ebird: A citizen-based bird observation network in the biological
  sciences.
\newblock {\em Biological conservation}, 2009.

\bibitem{thomee2016yfcc100m}
Bart Thomee, David~A Shamma, Gerald Friedland, Benjamin Elizalde, Karl Ni,
  Douglas Poland, Damian Borth, and Li-Jia Li.
\newblock Yfcc100m: The new data in multimedia research.
\newblock {\em Communications of the ACM}, 2016.

\bibitem{merlin_2020}
Cornell University.
\newblock Merlin bird id, 2020.
\newblock
  \url{https://apps.apple.com/us/app/merlin-bird-id-by-cornell-lab/id773457673}.

\bibitem{van2015building}
Grant Van~Horn, Steve Branson, Ryan Farrell, Scott Haber, Jessie Barry, Panos
  Ipeirotis, Pietro Perona, and Serge Belongie.
\newblock Building a bird recognition app and large scale dataset with citizen
  scientists: The fine print in fine-grained dataset collection.
\newblock In {\em CVPR}, 2015.

\bibitem{gvanhorn2018inat}
Grant Van~Horn, Oisin Mac~Aodha, Yang Song, Yin Cui, Chen Sun, Alex Shepard,
  Hartwig Adam, Pietro Perona, and Serge Belongie.
\newblock The inaturalist species classification and detection dataset.
\newblock In {\em CVPR}, 2018.

\bibitem{vedaldi2014understanding}
Andrea Vedaldi, Siddharth Mahendran, Stavros Tsogkas, Subhransu Maji, Ross
  Girshick, Juho Kannala, Esa Rahtu, Iasonas Kokkinos, Matthew Blaschko, David
  Weiss, et~al.
\newblock Understanding objects in detail with fine-grained attributes.
\newblock In {\em CVPR}, 2014.

\bibitem{wah2011caltech}
Catherine Wah, Steve Branson, Peter Welinder, Pietro Perona, and Serge
  Belongie.
\newblock The caltech-ucsd birds-200-2011 dataset.
\newblock 2011.

\bibitem{wu2018unsupervised}
Zhirong Wu, Yuanjun Xiong, Stella~X Yu, and Dahua Lin.
\newblock Unsupervised feature learning via non-parametric instance
  discrimination.
\newblock In {\em CVPR}, 2018.

\bibitem{yang2015large}
Linjie Yang, Ping Luo, Chen Change~Loy, and Xiaoou Tang.
\newblock A large-scale car dataset for fine-grained categorization and
  verification.
\newblock In {\em CVPR}, 2015.

\bibitem{yosinski2014transferable}
Jason Yosinski, Jeff Clune, Yoshua Bengio, and Hod Lipson.
\newblock How transferable are features in deep neural networks?
\newblock In {\em NeurIPS}, 2014.

\bibitem{zhai2019large}
Xiaohua Zhai, Joan Puigcerver, Alexander Kolesnikov, Pierre Ruyssen, Carlos
  Riquelme, Mario Lucic, Josip Djolonga, Andre~Susano Pinto, Maxim Neumann,
  Alexey Dosovitskiy, et~al.
\newblock A large-scale study of representation learning with the visual task
  adaptation benchmark.
\newblock {\em arXiv:1910.04867}, 2019.

\bibitem{zhang2016colorful}
Richard Zhang, Phillip Isola, and Alexei~A Efros.
\newblock Colorful image colorization.
\newblock In {\em ECCV}, 2016.

\bibitem{zhou2017places}
Bolei Zhou, Agata Lapedriza, Aditya Khosla, Aude Oliva, and Antonio Torralba.
\newblock Places: A 10 million image database for scene recognition.
\newblock {\em PAMI}, 2017.

\bibitem{pinkthroatedbrilliant}
T. Züchner, C.J. Sharpe, and P.F.D. Boesman.
\newblock Pink-throated brilliant (heliodoxa gularis).
\newblock {\em Birds of the World}, 2020.

\end{thebibliography}
}

\clearpage
\appendix
In this supplementary material we provide additional information related to how we constructed the iNat2021 dataset, statistics related to the distribution of images contained within, and provide detailed accuracy results. 

\section{iNat2021}
As with all dataset construction efforts, the devil is in the details regarding how data is sampled, the biases that are present as a result of those sampling techniques, and the overall resulting  quality. 
We try to be as clear as possible below regarding our construction steps so that researchers can understand and embrace these biases.

\subsection{Dataset Construction} 
The iNat2021 dataset was constructed by processing a database export from iNaturalist created on September 25th, 2020.
This export contains over 19M ``research grade'' observations across 200K species.
``Research grade'' refers to those observations where the iNaturalist community has reached a majority consensus on the species identification. 
The natural distribution of the observations in this export is extremely skewed toward a small number of common species \eg 50\% of the observations submitted to iNaturalist in the last year were classified to only 1\% of the total species observed in the last year. 
To construct a dataset amenable to computer vision experiments we had to make three primary decisions: (1) which species to include, (2) how to sample images for these species, and (3) how to split the images into train and test sets. To make these decisions we took the following steps, starting from the raw database export:
\begin{enumerate}[leftmargin=*]
    \itemsep-0.25em 
    \item We ``roll up'' all sub-species observations to their respective species node in the tree of life. This means that a species might have several different appearance clusters (\eg see Yellow-rumped Warbler\footnote{https://www.inaturalist.org/taxa/145245-Setophaga-coronata}).
    \item We restrict the candidate species to those in the kingdoms Animalia, Plantae and Fungi. This removes virus, mold species, and other species we deemed out of scope. 
    \item We restrict the candidate species to those that have at least 50 observations from 10 unique observers during the time period between 9/25/2019 - 9/25/2020. The idea behind this filter is to remove those species that were not observed ``enough'' times by ``enough'' people in the last year, where we arbitrarily decided the value of ``enough.'' Observations in this date range are eligible for the \textbf{test split} of the dataset. 
    \item We further restrict the candidate species to those that have at least 60 observations from before 9/25/2019. Note that we do not apply a unique observer constraint here. Observations in this date range are eligible for the \textbf{train} and \textbf{validation splits} of the dataset.
    \item 14k species passed the above filters. From this candidate pool, we kept the 10k species that have the most observations prior to 9/25/2019. The idea here is to keep the number of species in the dataset at a manageable size and to prefer those species that have been observed most. 
    \item For each of the selected species we then sample test images by iteratively choosing one observation (observed between 9/25/2019 - 9/25/2020) from each observer of a species until we have sampled 50 observations. This sampling protocol spreads the selected observations across observers in an attempt to mitigate the biases of any one observer (\eg camera type, location, composition preferences, etc.).
    \item Similarly, for the train and validation splits we iteratively choose one observation (observed before 9/25/2019) from each observer of a species until we have have sampled at least 60 and at most 310 observations. We split this collection into training and validation splits:
    \begin{itemize}
        \itemsep-0.25em 
        \item The validation set is taken to be 10 random observations for each species from this set, leaving at least 50 training observations for each species. 
        \item The full training set has at most 300 observations per species.
    \end{itemize}
\end{enumerate}

This results in 158873, 79172, 32594, and 94061 observers in the train, mini train, validation, and test splits. 
On average, these individuals observed 12.8, 6.3, 3.1, and 5.2 unique species each.

\subsection{Data Quality}
\label{sec:inat_data_quality}
Here we discuss some challenges related to data quality for the iNat2021 dataset, which are common to other large-scale natural world datasets. 
First, we are using the ``research grade'' labels from the iNaturalist community as the ground truth labels for the images. 
There is certainly a degree of error with these labels, a problem that afflicts most fine-grained classification datasets~\cite{van2015building}. 
In fact, an adhoc, small-scale, analysis of a small collection of ``research grade'' labels found the accuracy of the iNaturalist community to match an expert about 85\% of the time\footnote{iNaturalist Identification Quality,~\url{https://forum.inaturalist.org/t/identification-quality-on-inaturalist/7507}}. 
These analyses each have their own biases in terms of the observations analyzed (both location and taxa) and the criteria for an ``expert.'' It is therefore unclear whether the accuracy of the entire archive is closer to 95\% or 85\%.
We do not attempt to mitigate these errors with subsequent validation. Instead we move the burden of clean, high quality evaluation tasks to our NeWT dataset. 
However, it is important to note that from the perspective of self-supervised training, label noise should be irrelevant. 

Second, a research grade observation from iNaturalist is \textit{evidence} that a species was present at the photographed location. This means that the photograph might not actually contain the individual, but might be a footprint, scat, a nest or burrow, or even just markings on a tree (\eg beaver chew marks on a tree, or antler rubs from a deer). Similarly, the organism itself might be in one of several various life stages: alive vs dead, larval vs adult, breeding vs non-breeding, \etc{} Again, we embrace this data diversity and do not filter these images from either the train or test splits, nor do we attempt to control for the distribution of each type of modality within a particular species. 

Third, each image in the iNat2021 dataset has exactly one species label, even though in a small percentage of cases multiple species might be predominately featured in the photo. A classic example of this is an image of a bee pollinating a flower: is the correct classification the bee species or the flower species? Neither is incorrect. We do not attempt to mitigate this problem, and instead report multiple top-k classification results when benchmarking the dataset. Additionally, we attempt to address this nuisance with the introduction of our binary tasks in NeWT.

\subsection{Object Size Statistics}
\begin{figure}[t]
\includegraphics[width=0.45\textwidth]{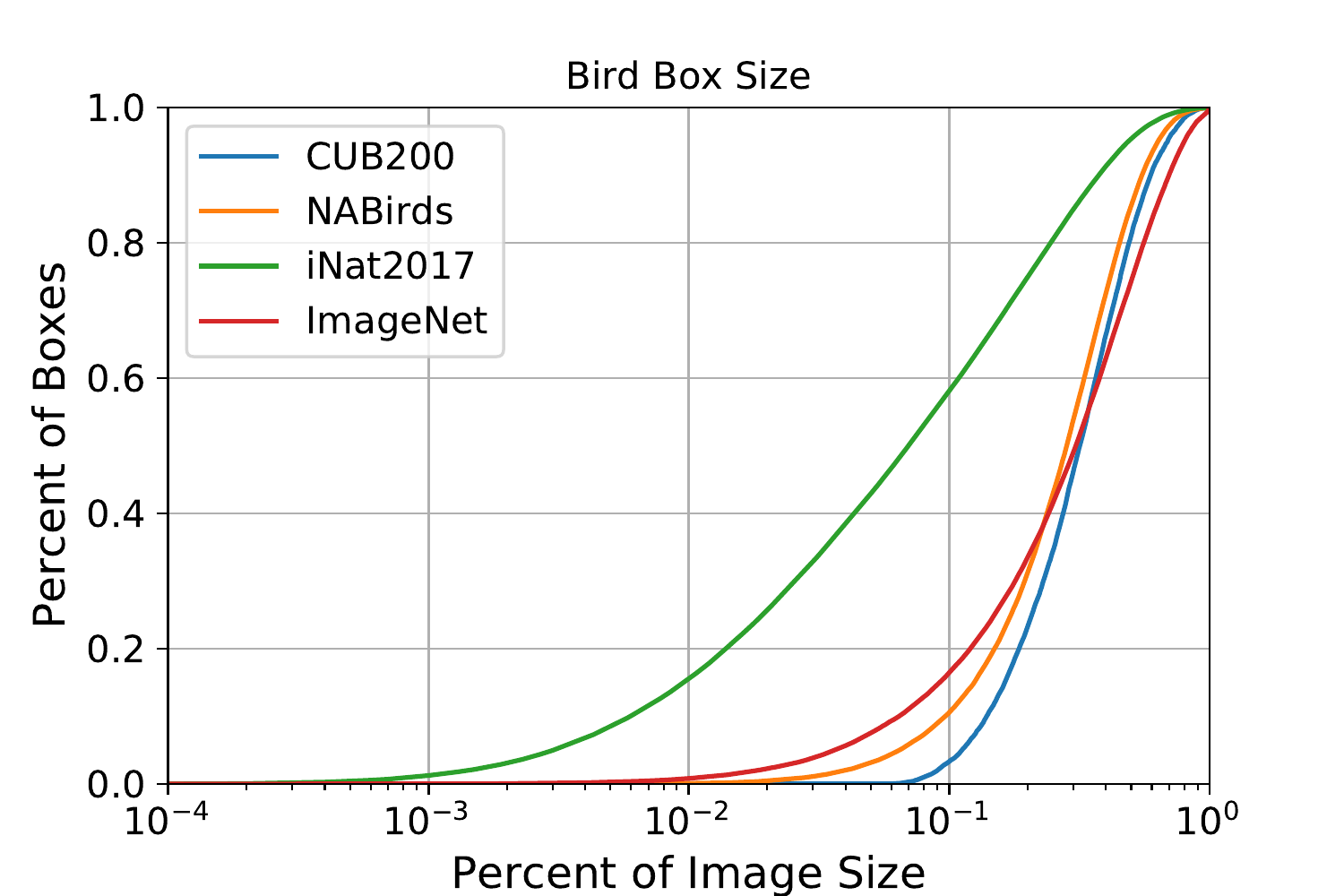}
\caption{Cumulative fraction of ground truth bird bounding boxes as a fraction of image size. We can see that the birds found in CUB, NABirds, and ImageNet are all relatively large, with $\sim{}50\%$ of bird boxes taking up at least $\sim{}40\%$ of the image size. However, the distribution of bird boxes in iNat2017, which should be representative of the birds in iNat2021, are much smaller, with $\sim{}50\%$ of the boxes having an area less than $\sim{}7\%$ of the image size. This discrepancy in bird sizes might explain why ImageNet is actually a better pretraining dataset for CUB and NABirds than iNat2021 is (when using self-supervised techniques designed for ImageNet).}
\label{fig:bird_size_comparison}
\end{figure}

In Figure~\ref{fig:bird_size_comparison} we plot the cumulative fraction of ground truth bird bounding boxes from CUB, NABirds, iNat2017, and ImageNet as a fraction of the image size. The iNat2017 dataset~\cite{gvanhorn2018inat} had manually provided bounding boxes for the bird species, and these boxes should also be representative of the birds in the iNat2021 dataset. We can see that the distribution of bounding boxes for birds in CUB, NABirds, and ImageNet are very similar, with 50\% of the bird bounding boxes having an area \textit{greater than} 40\% of the image size. Meanwhile, the birds in iNat2017 are much smaller, with 50\% of the birds having an area \textit{less than} 7\% of the image size. Self-supervised techniques like SimCLR were designed for ImageNet, therefore their default augmentation strategy has been designed for objects that take up a large fraction of the image size. Applying these strategies to datasets where objects do not take a large fraction of the image size may be inappropriate (see the performance of self-supervised iNat2021 mini in the main paper). 
Note that supervised methods can still recover discriminative features (see the performance of supervised iNat2021 mini in the main paper), so there is hope that self-supervised methods can also work on these datasets.

\subsection{iNat2021 Geographical Distribution}

\begin{figure*}[t]
\setlength{\tabcolsep}{2pt}
\renewcommand{\arraystretch}{1.0}
\begin{tabular}{ccc}
\centering
\includegraphics[width=0.33\linewidth]{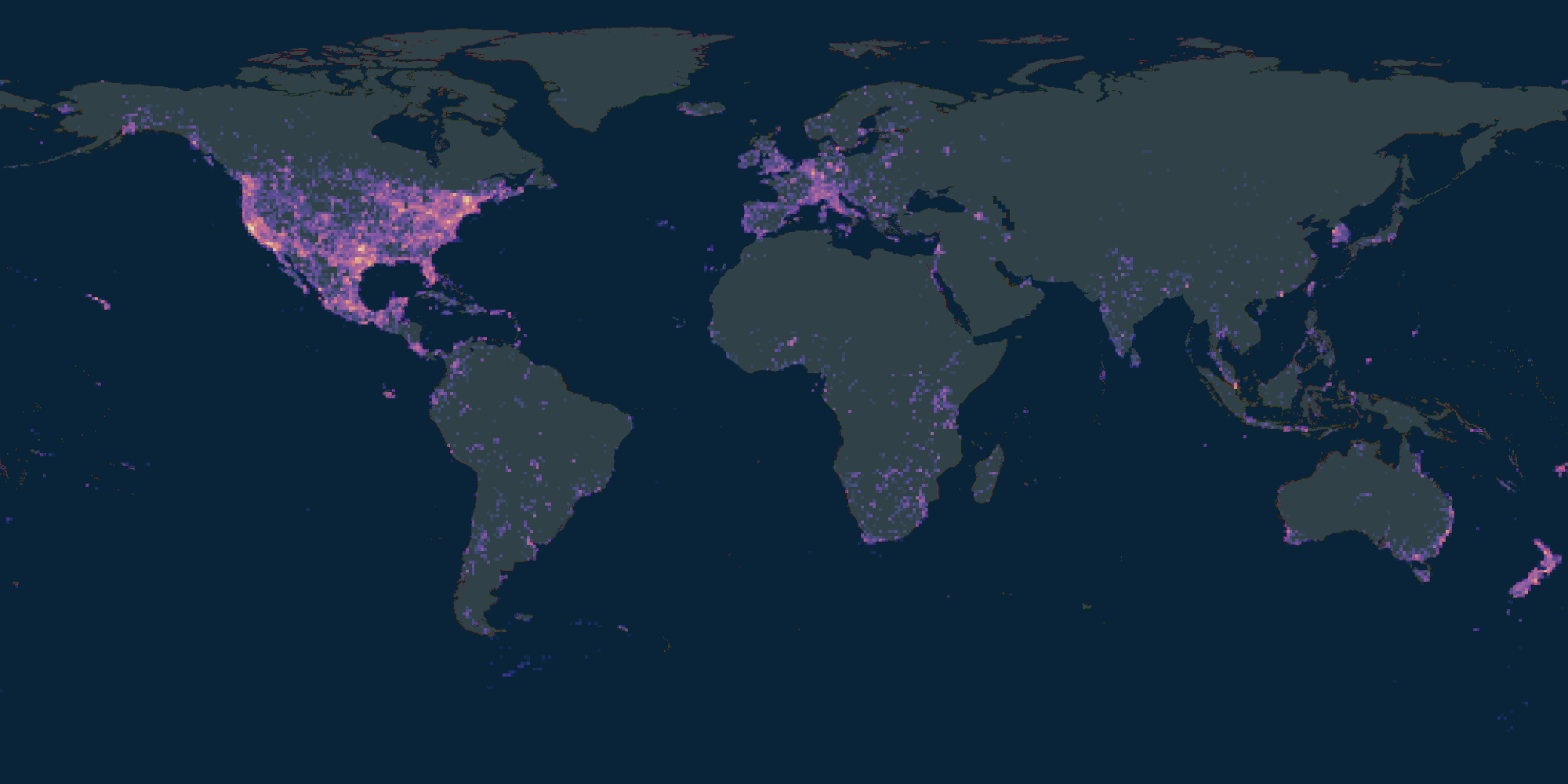}
&
\includegraphics[width=0.33\linewidth]{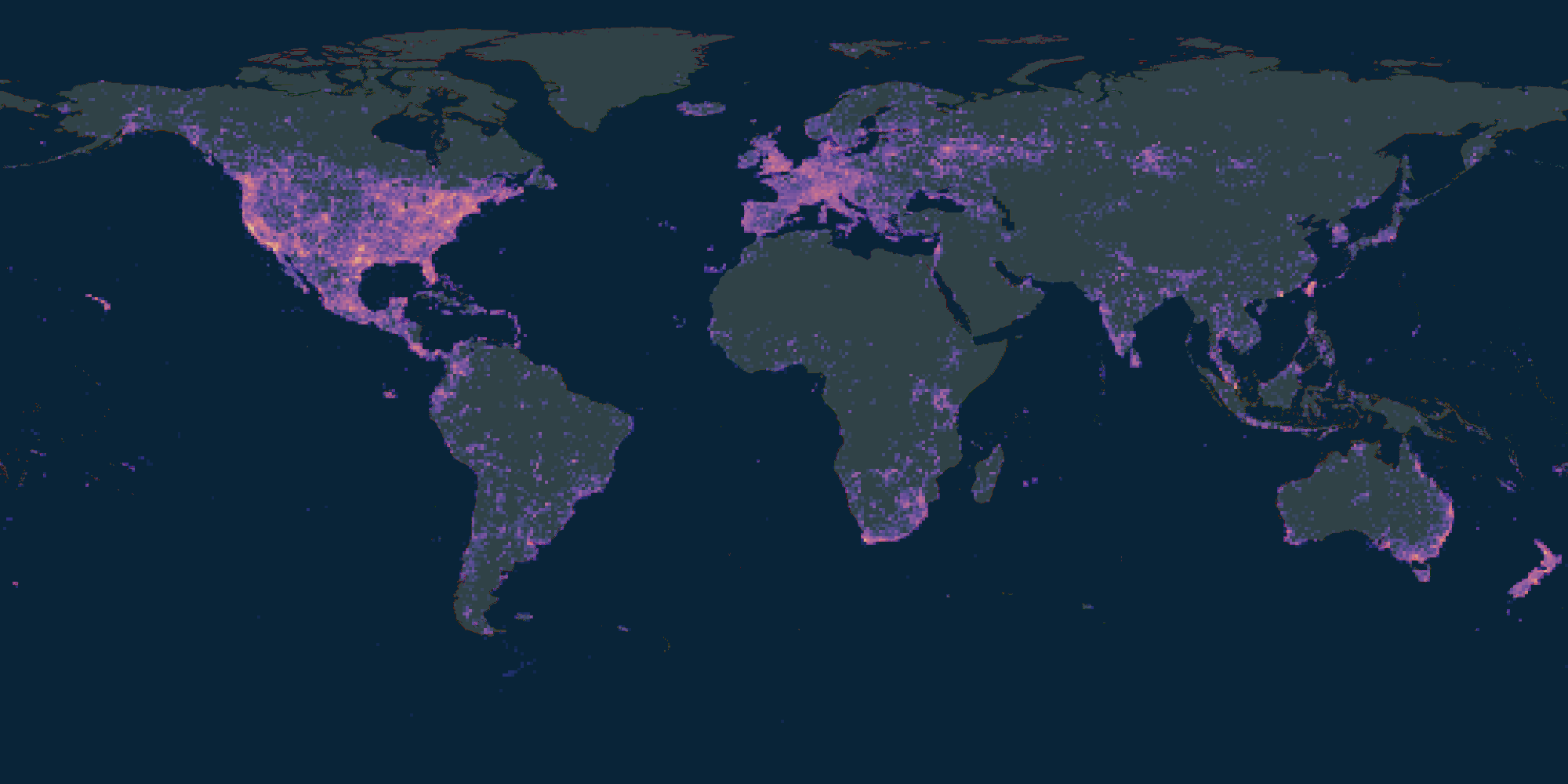}
&
\includegraphics[width=0.33\linewidth]{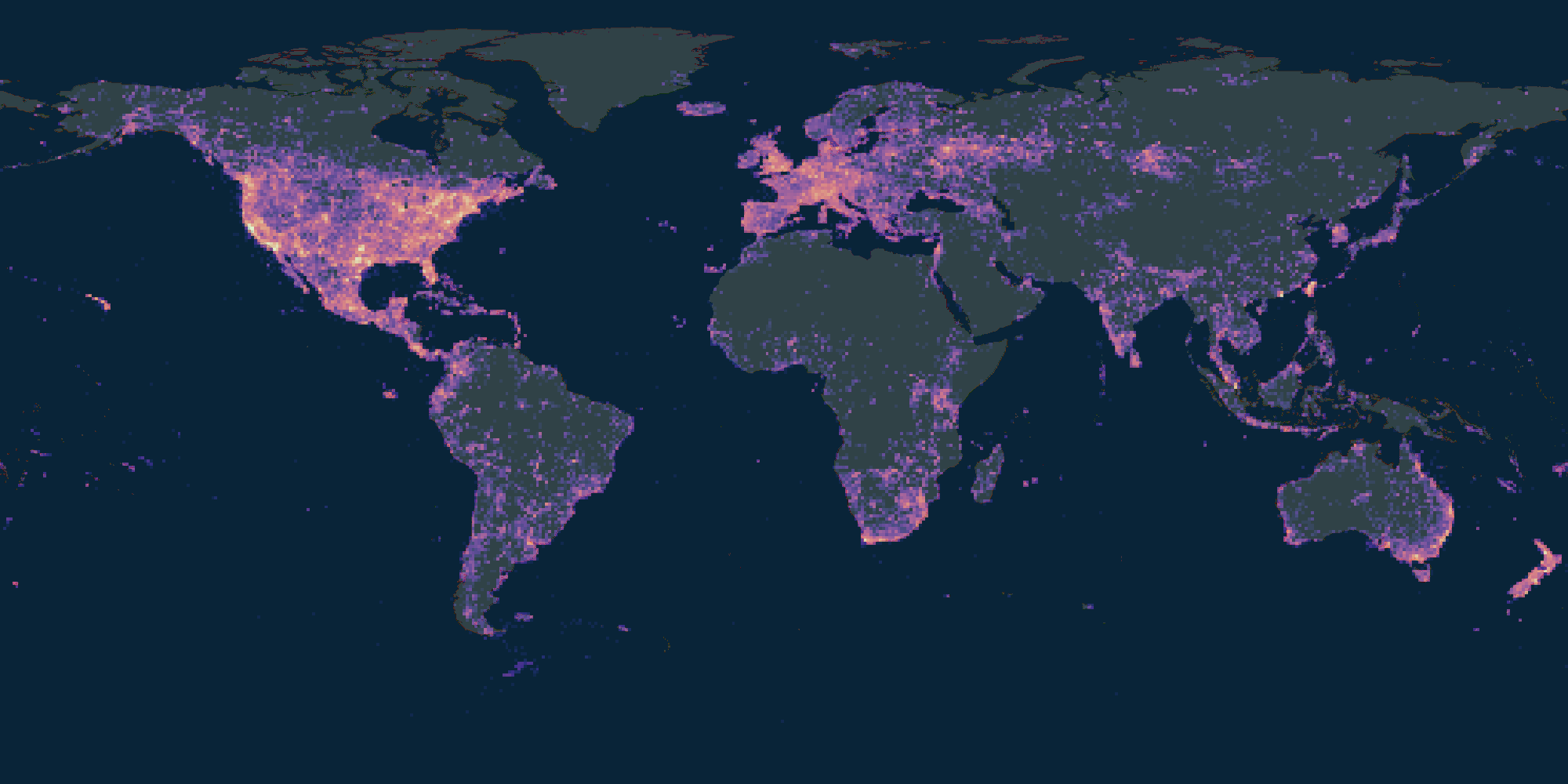}
\\
iNat2018 & iNat2021 mini & iNat2021

\end{tabular}
\caption{Histogram of the locations of training images.  By comparing our new iNat2021 dataset to the 2018 variant, we can see that iNat2021 is more geographically varied. 
Not visualized here, but we observe no major difference between the distribution of the training and test images for iNat2021, despite the test images being sampled between September 2019 to September 2020.  
}
\label{fig:map}
\end{figure*}

% \begin{figure*}[t]
% \centering
% \begin{subfigure}[h]{0.32\linewidth}
% \includegraphics[width=\linewidth]{figures/maps/map_2018.png}
% \caption{iNat2018}
% \end{subfigure}
% \begin{subfigure}[h]{0.32\linewidth}
% \includegraphics[width=\linewidth]{figures/maps/map_2020_mini.png}
% \caption{iNat2021 mini}
% \end{subfigure}
% \begin{subfigure}[h]{0.32\linewidth}
% \includegraphics[width=\linewidth]{figures/maps/map_2020_large.png}
% \caption{iNat2021}
% \end{subfigure}%
% \caption{Histogram of the locations of training images.  By comparing our new iNat2021 dataset to other related ones, we can see that iNat2021 is more geographically varied. 
% Not visualized here, but we observe no discernible difference between the distribution of the training and test images for iNat2021, despite the test images being sampled between September 2019 to September 2020.  
% }
% \label{fig:map}
% \end{figure*}

The overwhelming majority of images in iNat2021 also include metadata indicating when and where they were captured. 
In Figure~\ref{fig:map} we illustrate the geographical distribution of the training images. 
Not only does the dataset contain more images, it is also more geographically diverse, consisting of more images from a larger number of countries. 
This partially addresses the North American bias of existing datasets such as iNat2018 and enables more realistic exploration of the impact of diverse geographical distributions for problems like federated learning.%~\cite{hsu2020federated}. 

To further explore the value of the image location information in iNat2021 we train a geographical prior from \cite{mac2019presence} and combine it with the predictions of our supervised image classifiers. 
As in \cite{mac2019presence}, we train a fully connected neural network that takes location information as input and predicts a distribution over the classes that are likely to be present at the specified location. 
We do not use time or photographer information during training, and we do not artificially balance the classes during each epoch. 
Otherwise, we use the default settings as the public code. 
The results are presented in Table \ref{table:geo_prior_results}. 
As expected, we see a noticeable improvement when we combine the output of any geographical prior with the image classifier (here a supervised ResNet50). 
We also see that combining the geographical prior that has been trained on the full dataset with the vision model that has been trained on the mini set is superior to only using the mini geographical prior. 
This highlights the complementary value of the different signals.

\begin{table}[h]
\small
\centering
\begin{tabular}{|l l | c c|} 
 \hline
 Train Vision  & Train Geo & Val Acc & Test Acc \\ 
 \hline\hline
 iNat2021 mini & None & 0.658 & 0.654 \\ 
 iNat2021 & None & 0.764 &  0.760 \\ \hline
 iNat2021 mini & iNat2021 mini & 0.719 & 0.722  \\ 
 iNat2021 mini & iNat2021 & 0.739 & 0.739 \\ 
 iNat2021 & iNat2021 mini & 0.781 & 0.785 \\ 
 iNat2021 & iNat2021 & 0.801 & 0.800 \\ 
 \hline
\end{tabular}
\vspace{5pt}
\caption{Top-1 classification accuracy after combing a supervised image classifier (here a ResNet50) with the learned geographical prior from \cite{mac2019presence}, retrained on iNat2021. The `Train Vision' and `Train Geo' columns indicate which dataset the respective models have been trained on.
}
\label{table:geo_prior_results}
\end{table}

\section{Additional Results} 

%\subsection{Detailed FG Dataset Results}
See Table~\ref{tab:fg_detailed_accuracy} for the top-1 accuracy values achieved by each method on the various fine-grained datasets.
%\subsection{Detailed NeWT Results}
See Table~\ref{tab:newt_detailed_accuracy} for the top-1 accuracy values achieved by each method on the various NeWT tasks.

\begin{table*}[h]
\footnotesize
\centering
\begin{tabular}{|l | l | c c c c c c | c |}
\hline
Source Dataset  &  	Train Loss  &  	\vtop{\hbox{\strut Oxford}\hbox{\strut Flowers}}  &  	CUB  &  	CUBExpert  &  	NABirds  &  	\vtop{\hbox{\strut Stanford}\hbox{\strut Dogs}}  &  	\vtop{\hbox{\strut Stanford}\hbox{\strut Cars}}  &  	\vtop{\hbox{\strut Mean}\hbox{\strut ACC}}\\
\hline\hline
ImageNet  &  	Supervised  &  	0.820  &  	0.650  &  	0.669  &  	0.542  &  	0.890  &  	0.479  &  	0.675 \\
ImageNet  &  	SimCLR  &  	0.798  &  	0.445  &  	0.452  &  	0.332  &  	0.587  &  	0.425  &  	0.507 \\
ImageNet  &  	SimCLR x4  &  	0.850  &  	0.535  &  	0.550  &  	0.433  &  	0.722  &  	0.497  &  	0.598 \\
ImageNet  &  	SimCLR v2  &  	0.839  &  	0.552  &  	0.565  &  	0.416  &  	0.629  &  	0.476  &  	0.580 \\
ImageNet  &  	SwAV  &  	0.877  &  	0.631  &  	0.647  &  	0.534  &  	0.727  &  	0.595  &  	0.668 \\
ImageNet  &  	MoCo v2  &  	0.824  &  	0.476  &  	0.488  &  	0.377  &  	0.630  &  	0.437  &  	0.539 \\ \hline
iNat2021  &  	Supervised (from ImageNet)  &  	0.937  &  	0.864  &  	0.891  &  	0.804  &  	0.539  &  	0.177  &  	0.702 \\
iNat2021  &  	Supervised  &  	0.927  &  	0.862  &  	0.889  &  	0.797  &  	0.522  &  	0.161  &  	0.693 \\
iNat2021 mini  &  	Supervised (from ImageNet)  &  	0.904  &  	0.841  &  	0.866  &  	0.763  &  	0.564  &  	0.191  &  	0.688 \\
iNat2021 mini  &  	Supervised  &  	0.886  &  	0.820  &  	0.842  &  	0.726  &  	0.474  &  	0.211  &  	0.660 \\
iNat2018  &  	Supervised (from ImageNet)  &  	0.867  &  	0.862  &  	0.891  &  	0.806  &  	0.554  &  	0.203  &  	0.697 \\ \hline
iNat2021  &  	SimCLR  &  	0.863  &  	0.365  &  	0.371  &  	0.267  &  	0.391  &  	0.252  &  	0.418 \\
iNat2021 mini  &  	SimCLR  &  	0.814  &  	0.308  &  	0.316  &  	0.224  &  	0.366  &  	0.295  &  	0.387 \\
iNat2021 mini  &  	SimCLR x4  &  	0.911  &  	0.420  &  	0.432  &  	0.328  &  	0.452  &  	0.376  &  	0.486 \\
iNat2021 mini  &  	SimCLR v2  &  	0.859  &  	0.361  &  	0.367  &  	0.260  &  	0.393  &  	0.274  &  	0.419 \\
iNat2021 mini  &  	SwAV  &  	0.795  &  	0.256  &  	0.258  &  	0.169  &  	0.342  &  	0.225  &  	0.341 \\
iNat2021 mini  &  	MoCo v2  &  	0.825  &  	0.335  &  	0.349  &  	0.253  &  	0.390  &  	0.257  &  	0.401 \\
\hline
\end{tabular}
\vspace{5pt}
\caption{Top-1 classification accuracy for ResNet50 pretrained on different source datasets and used as a fixed feature extractor on each respective fine-grained dataset.}
\label{tab:fg_detailed_accuracy}
\end{table*}

\begin{table*}[h]
\footnotesize
\centering
\begin{tabular}{|l | l | c c c c c c c c | c |}
\hline
Source Dataset  &  	Train Loss  &  	\vtop{\hbox{\strut Appear.}\hbox{\strut Age}}   &  	\vtop{\hbox{\strut Appear.}\hbox{\strut Attribute}}  &  	\vtop{\hbox{\strut Appear.}\hbox{\strut Health}}  &  	\vtop{\hbox{\strut Appear.}\hbox{\strut Species}}  &  	Behavior  &  	Context  &  	Counting  &  	Gestalt  &  	\vtop{\hbox{\strut Mean}\hbox{\strut ACC}}\\
\hline\hline
ImageNet  &  	Supervised  &  	0.793  &  	0.846  &  	0.841  &  	0.725  &  	0.715  &  	0.813  &  	0.595  &  	0.859  &  	0.744 \\
ImageNet  &  	SimCLR  &  	0.702  &  	0.732  &  	0.807  &  	0.647  &  	0.675  &  	0.792  &  	0.590  &  	0.891  &  	0.678 \\
ImageNet  &  	SimCLR x4  &  	0.718  &  	0.724  &  	0.806  &  	0.652  &  	0.687  &  	0.829  &  	0.600  &  	0.899  &  	0.684 \\
ImageNet  &  	SimCLR v2  &  	0.770  &  	0.751  &  	0.819  &  	0.671  &  	0.707  &  	0.812  &  	0.550  &  	0.897  &  	0.705 \\
ImageNet  &  	SwAV  &  	0.813  &  	0.793  &  	0.824  &  	0.709  &  	0.715  &  	0.831  &  	0.590  &  	0.902  &  	0.733 \\
ImageNet  &  	MoCo v2  &  	0.743  &  	0.758  &  	0.831  &  	0.683  &  	0.692  &  	0.795  &  	0.580  &  	0.892  &  	0.709 \\ \hline
iNat2021  &  	Supervised (from ImageNet)  &  	0.894  &  	0.872  &  	0.854  &  	0.814  &  	0.761  &  	0.764  &  	0.620  &  	0.853  &  	0.806 \\
iNat2021  &  	Supervised  &  	0.893  &  	0.854  &  	0.864  &  	0.816  &  	0.755  &  	0.759  &  	0.590  &  	0.870  &  	0.806 \\
iNat2021 mini  &  	Supervised (from ImageNet)  &  	0.879  &  	0.842  &  	0.845  &  	0.803  &  	0.740  &  	0.760  &  	0.590  &  	0.864  &  	0.793 \\
iNat2021 mini  &  	Supervised  &  	0.872  &  	0.865  &  	0.843  &  	0.789  &  	0.721  &  	0.744  &  	0.610  &  	0.852  &  	0.782 \\
iNat2018  &  	Supervised (from ImageNet)  &  	0.861  &  	0.850  &  	0.848  &  	0.804  &  	0.719  &  	0.774  &  	0.605  &  	0.877  &  	0.791 \\ \hline
iNat2021  &  	SimCLR  &  	0.688  &  	0.723  &  	0.823  &  	0.661  &  	0.674  &  	0.822  &  	0.640  &  	0.900  &  	0.689 \\
iNat2021 mini  &  	SimCLR  &  	0.707  &  	0.710  &  	0.797  &  	0.648  &  	0.647  &  	0.792  &  	0.565  &  	0.888  &  	0.678 \\
iNat2021 mini  &  	SimCLR x4  &  	0.700  &  	0.718  &  	0.837  &  	0.662  &  	0.682  &  	0.829  &  	0.565  &  	0.908  &  	0.689 \\
iNat2021 mini  &  	SimCLR v2  &  	0.702  &  	0.719  &  	0.818  &  	0.660  &  	0.667  &  	0.807  &  	0.560  &  	0.903  &  	0.687 \\
iNat2021 mini  &  	SwAV  &  	0.625  &  	0.677  &  	0.797  &  	0.654  &  	0.644  &  	0.822  &  	0.545  &  	0.881  &  	0.675 \\
iNat2021 mini  &  	MoCo v2  &  	0.709  &  	0.735  &  	0.830  &  	0.664  &  	0.685  &  	0.795  &  	0.570  &  	0.875  &  	0.691 \\
\hline
\end{tabular}
\vspace{5pt}
\caption{Top-1 classification accuracy for ResNet50 pretrained on different source datasets and used as a fixed feature extractor on NeWT. The mean accuracy column is computed by averaging the top-1 accuracy across all tasks.}
\label{tab:newt_detailed_accuracy}
\end{table*}

\section{Additional Implementation Details} 
For all methods we used the code repositories provided by the respective authors. For models self-supervised on ImageNet, we used the checkpoints that were available in the code repositories. For models self-supervised on iNat2021 we used default parameters unless otherwise stated. For the sake of time and cost, we only train on iNat2021 mini for these experiments:\\
\textbf{SimCLR V1 ResNet50 4x ~\cite{chen2020simple}}: We train a ResNet50 with a width multiplier of 4. This is a very large model with over 375M parameters in the backbone. The model is trained to 1000 epochs using a batch size of 1024. Note the fixed feature representation size is 8192 as opposed to 2048. \\
\textbf{SimCLR V2~\cite{chen2020big}}: We train a ResNet50 model using a 3 layer projection head. We do not incorporate the memory mechanism from MoCo, and we do not do any distillation. We completely throw away the project head and extract features using only the backbone. The model is trained to 1000 epochs using a batch size of 1024.\\
\textbf{SwAV~\cite{swav_2020}}: We train a ResNet50 model using a batch size of 256 for a total of 200 epochs. We set epsilon to 0.03 and started the queue at epoch 30. \\
\textbf{MoCo V2~\cite{chen2020mocov2}}: We train a ResNet50 model using a batch size of 256 for a total of 1000 epochs. 

The fine-grained datasets experiments used a linear SVM model trained using SGD, while the NeWT experiments used a linear SVM model trained using liblinear. This difference was motivated by convenience: the NeWT tasks are all small enough ($\le$200 training features) where liblinear can be used effectively. We found that the  fine-grained datasets (with thousands of training features) were too large for liblinear and therefore we resorted to SGD.

\section{Additional NeWT Information}

\begin{figure*}[h]
\centering
\includegraphics[width=\textwidth]{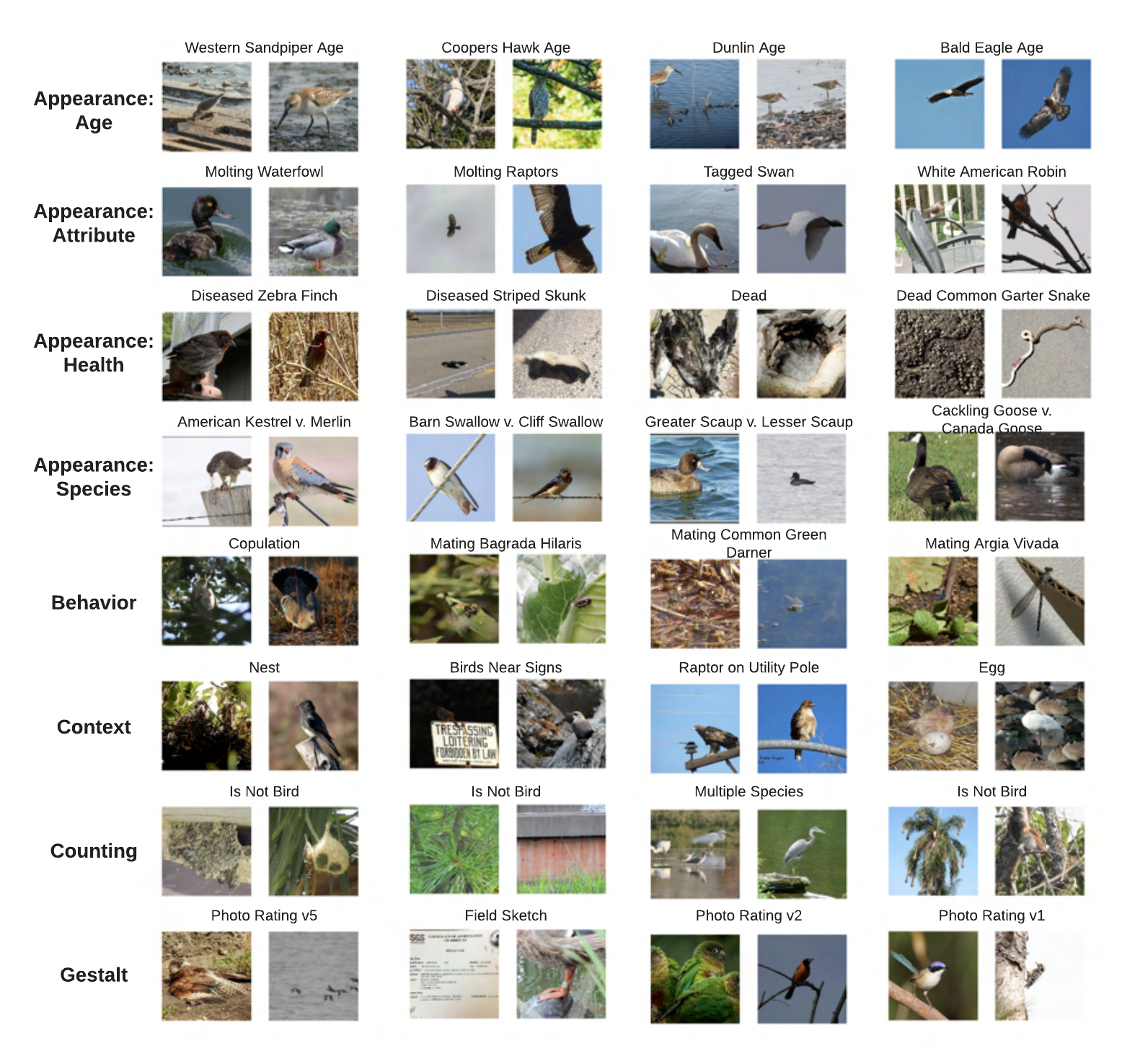}
\caption{Example pairs of tasks from each group. For each pairs of images, representative positive instances are on the left and negatives are on the right.}
    \label{fig:task_viz}
\end{figure*}

In Figure~\ref{fig:task_viz} we visualize some examples from a subset of tasks from NeWT. Note the images have been center cropped and reduced in size.

\end{document}